%% file: main.tex
\documentclass{article} 
\usepackage[T1]{fontenc}
\usepackage{iclr2027_conference,times}

\input{math_commands.tex}

\usepackage{float}
\usepackage{hyperref}
\usepackage{url}
\usepackage{amsmath,amssymb}
\usepackage{booktabs}
\usepackage{multirow}
\usepackage{graphicx}
\usepackage{xspace}
\usepackage{microtype}
\usepackage{tikz}

\usetikzlibrary{arrows.meta,positioning,calc,decorations.pathreplacing,fit,backgrounds}
\definecolor{cState}{RGB}{225,236,250}\definecolor{cStateLine}{RGB}{54,98,160}\definecolor{cWriter}{RGB}{255,224,178}\definecolor{cWriterLine}{RGB}{190,110,20}
\definecolor{cFact}{RGB}{190,30,45}\definecolor{cFactFill}{RGB}{252,228,230}\definecolor{cDec}{RGB}{30,110,60}\definecolor{cDecFill}{RGB}{226,244,232}\definecolor{cGray}{RGB}{150,150,150}
\graphicspath{{figures/}}

\newcommand{\Fp}{F'}                       
\newcommand{\zrt}{z^{\mathrm{rt}}}         
\newcommand{\NLL}{\mathrm{NLL}}
\newcommand{\SK}{S_K}
\newcommand{\dssr}{DSSR\xspace}
\newcommand{\fmt}[1]{\textsc{#1}}          
\newcommand{\ci}[2]{[#1,\,#2]}             
\newcommand{\pts}{\,pts}

\title{Decision-Sufficient State Representations: Measuring and Reducing Write-Time Regret}

\author{Bingyu Shen \\
Independent Researcher \\
\texttt{bingyu.shen@hotmail.com}
\And
Boyang Li \\
Department of Computer Science and Technology \\
Kean University, Union, NJ, USA \\
\texttt{boli@kean.edu}
}

\iclrfinalcopy 
\begin{document}

\thanks{Preprint. Under review.}

\maketitle
\lhead{Preprint}        

\begin{abstract}
\input{0_abs}
\end{abstract}

\input{1_intro}

\section{Related work}
\label{sec:related}

\textbf{Memory and state for LLM agents:}
Recursive summarisation rewrites a running summary as the interaction proceeds \citep{wang2023recursive}; MemGPT pages information between context and external storage \citep{packer2023memgpt}; Reflexion, generative agents, MemoryBank and A-MEM keep reflections and retrievable memories \citep{shinn2023reflexion,park2023generative,zhong2024memorybank,xu2025amem}. Closer to our writer are compact-state designs: belief claims with verbalised uncertainty \citep{singh2026agentbrace}, progress-aware retention \citep{jiang2026pabu}, competing conclusions with probabilities \citep{liao2026beliefmem}, optimised compression guidelines \citep{kang2025acon}, context folding \citep{sun2025contextfolding}, and MEM1's end-to-end reinforcement learning of a constant-memory agent \citep{zhou2025mem1}. They propose memory designs; we measure where a fixed design loses and train the writer alone from a frozen reader's loss. Our prompted formats (\S\ref{sec:setup}) mirror this literature: a free-text summary, a belief list with certainty tags (Agent-BRACE-like), fixed slots (PABU-like), and a summary with a hand-written retention guideline of the kind ACON optimises.

\textbf{Context compression and evaluation:}
Gist tokens, in-context autoencoders, AutoCompressors and LLMLingua compress a \emph{given} context at read time \citep{mu2023gist,ge2024icae,chevalier2023autocompressors,jiang2023llmlingua}; our read-time oracle $\zrt$ is such a compression chosen with hindsight, and write-time regret measures how far a prospective writer falls short of it. Long contexts do not remove the problem, since models use them unevenly \citep{liu2024lost}. LoCoMo, LongMemEval and MemoryAgentBench measure the end accuracy of memory systems over long interactions \citep{maharana2024locomo,wu2025longmemeval,hu2026memoryagentbench}; our protocol adds a decomposition of the loss into budget and writer terms and a lag manipulation verified to create lag.

\textbf{Hindsight and preference training:}
Hindsight relabeling turns achieved outcomes into supervision \citep{andrychowicz2017her,zhang2023wisdom}. \dssr relabels a write with the reader's future loss and trains the writer with direct preference optimization \citep{rafailov2023dpo}, adding a likelihood term on the chosen sample as in iterative reasoning preference optimization \citep{pang2024iterative}. The readers are behavior-cloned with DAgger iterations \citep{ross2011dagger}.

\section{Constant-context agents and write-time regret}
\label{sec:setting}

\textbf{The constant-context agent:}
A writer $U$ maintains a text state $z_t$ of at most $B$ tokens, updated recursively from the goal $g$, the previous state, the last action, and the new observation, $z_t = U(g, z_{t-1}, a_{t-1}, o_t)$; the raw history is discarded after each write. A frozen reader $\pi$ chooses the action from $(g, z_t, o_t)$ alone. Nothing but $z_t$ carries information across steps: it plays the role of a belief state \citep{kaelbling1998planning}, and the design has constant context by construction. We study the writer: the reader is trained once, on states from untrained writers, and never changes.

\textbf{Decision loss and its decomposition:}
Consider a decision point $t$ with reference action $a^*_t$ (the first command of the environment's optimal policy) and full history $h_t$. For any context $c$ let $\NLL_t(c) = -\log p_\pi(a^*_t \mid g, c, o_t)$, with $p_\pi$ the reader's distribution over the admissible options. The \emph{sufficiency gap} of a state format $M$ is $\delta_t(M) = \NLL_t(z^M_t) - \NLL_t(h_t)$. To separate budget from writer we define a \emph{read-time oracle} state $\zrt_t$: the untrained writer compresses the entire history $h_t$ in one shot at budget $B$, $N=8$ samples are drawn, and the one with the lowest $\NLL_t$ is kept. $\zrt_t$ is a hindsight construction, since it sees the whole history and is selected with $a^*_t$; 
it is a hindsight reference for what $B$ tokens written by this writer can carry for this decision, not a bound: a recursively written state can do better at a given decision.
Then
\begin{equation}
\delta_t = \underbrace{\NLL_t(\zrt_t) - \NLL_t(h_t)}_{\beta_t\ \text{(budget loss)}} \;+\; \underbrace{\NLL_t(z^M_t) - \NLL_t(\zrt_t)}_{\kappa_t(M)\ \text{(write-time regret)}}.
\label{eq:decomp}
\end{equation}
$\beta$ is what a $B$-token state must lose even with hindsight; $\kappa$ is what the recursive writer lost on top of that by choosing, prospectively, what to keep.

\textbf{Relevance lag:}
To relate regret to time we measure, for each decision point, which past step matters: the \emph{relevance lag} $\ell_t = t - s^*$, where $s^*$ is the chunk of $h_t$ whose removal increases $\NLL_t(h_t)$ most (chunks are defined in Appendix~\ref{app:metrics}), undefined when no removal costs at least 0.5 nats; $\ell$ is measured on the actual history, whatever the protocol intended.

\textbf{Controlled lag protocol:}
Episodes have two phases. In Phase A a scripted explorer visits every room, opening every container and door, so that every fact a later decision needs (ingredient locations, the knife, the appliances, the map) has been observed; it then walks to the kitchen and reads the cookbook, which reveals the recipe. In Phase B the reader acts until the game is won or lost. The writer updates $z_t$ at every step of both phases. Lag is manipulated by a seeded random walk of $\Fp \in \{0,10,20,40\}$ steps among visited rooms, placed \emph{after} the reveal (protocol $\Fp$); the walks for different $\Fp$ are nested prefixes of one walk. Our pre-registered design placed the walk \emph{before} the reveal and did not create lag (\S\ref{sec:diag}); protocol $\Fp$ was adopted before the writer was trained and before the test split was opened (original-protocol results in Appendix~\ref{app:stage2}).

\section{\dssr: training the writer from the reader's loss}
\label{sec:method}

\textbf{From episodes to single writes:}
Training the writer on episode outcomes would mean multi-turn reinforcement learning through a frozen reader with a sparse, delayed reward. \dssr instead decomposes the writer's problem into single-turn write problems. At a write point $t$, sibling candidate states $z^{(1)}, \dots, z^{(n)}$ sampled from the current writer are compared by how well the frozen reader can make the \emph{future} decisions from them, and the writer is trained to prefer the better candidates. The signal is the quantity we want to reduce, the reader's loss.

\textbf{Hindsight sufficiency score:}
For write point $t$ and candidate $z$, the targets are the decision steps $u$ of Phase B with $t < u \le t+40$, evenly subsampled to at most $K$ (default $K=12$). The score is the mean reader log-likelihood of the reference actions at the targets,
\begin{equation}
\SK(z) = \frac{1}{|T_K|} \sum_{u \in T_K} \log p_\pi\big(a^*_u \mid g,\; c_u(z),\; o_u\big),
\label{eq:score}
\end{equation}
where $c_u(z)$ is the context the reader sees at $u$. Two forms of $c_u$ are possible. The \emph{window} form is $z$ followed by the raw steps $t{+}1, \dots, u{-}1$: it credits $z$ only for information from before $t$ and is cheap, and it is the form most hindsight schemes would use. The \emph{recursive} form rolls the writer forward from $z$ through the logged steps, $z_k = U(g, z_{k-1}, a_{k-1}, o_k)$ at temperature 0, and the reader sees only $z_u$, exactly as it would online. The two forms disagree completely for a lossy writer (\S\ref{sec:diag}): the value of a state depends on how it will be rewritten, and the window form hides this. \dssr uses the recursive form.

\textbf{Optimization:}
Each round collects episodes on the training games with the current writer and the frozen reader, samples write points from both phases, writes $n$ candidates per point at temperature 1, scores them with $\SK$, and trains. Round 0 is supervised fine-tuning on the best candidate by $S_{12}$ (best-of-$n$ distillation, as in reward-ranked fine-tuning, \citealp{dong2023raft}). Rounds 1--3 are direct preference optimization \citep{rafailov2023dpo} on best-versus-worst pairs whose score gap is at least 0.3 nats (at most two pairs per write point), with $\beta_{\mathrm{DPO}}=0.1$ and a 0.5-weighted likelihood term on the chosen state \citep{pang2024iterative}; the previous round's adapter is both the initialisation and the reference. The writer is a LoRA adapter \citep{hu2022lora}; the prompt is the unchanged summary prompt, so \dssr and the \fmt{summary} baseline differ only in the adapter. Model selection uses online success on the validation games, and the selected configuration is retrained with three seeds. Variants trained from the same candidates with the same recipe isolate the objective: horizon $K \in \{1, 4, 12\}$; \fmt{no-hint}, which rejects candidates that contain an admissible command verbatim; and \fmt{outcome}, which scores a candidate by the true return of continuing the episode from it, the outcome-reward alternative.

\textbf{Reachability-restricted sampling:}
Section~\ref{sec:credit} shows that under lag most write points offer no candidate that still contains the decisive fact. The \emph{A4} variant keeps a post-reveal write point only if its input state still holds at least one recipe directive, so that keeping versus dropping is a choice the candidates can make, and writes 12 candidates at temperature 1.1 so that both choices appear. Everything else is unchanged. This variant produced the selected writer.

\section{Experiments}
\label{sec:exp}

\subsection{Setup}
\label{sec:setup}

\textbf{Environment and splits:}
We use TextWorld cooking games \citep{cote2018textworld,adolphs2020ledeepchef,adhikari2020gata} with three-ingredient recipes, cutting and cooking, and 6, 9 or 12 rooms (240 train, 45 validation and 90 test games, disjoint seeds; the validation and test games contain unseen foods and preparations). A game is won by preparing and eating the meal and is lost, irreversibly, on the first wrong cut or cooking verb. Full histories run from 2.2k tokens (6 rooms, $\Fp=0$) to 9.1k on average (12 rooms, $\Fp=40$), against state budgets $B \in \{64, 128, 256\}$. The reader chooses among the admissible commands, presented as lettered options and shuffled with a fixed seed; the reference action $a^*$ is the first command of the game's optimal policy, which is privileged and never shown to the writer or reader (a string search over all logged prompts enforces this). Details and environment statistics are in Appendix~\ref{app:env}.

\textbf{Models and formats:}
Writer and reader $R_1$ are Qwen3-4B-Instruct \citep{yang2025qwen3} with separate LoRA adapters served by vLLM \citep{kwon2023vllm} with batch-invariant kernels, so every number is bit-reproducible on one GPU. $R_1$ is behaviour-cloned on the reference action over 19 context variants (the full history and each prompted format at each budget, from \emph{untrained} writers), with two DAgger iterations \citep{ross2011dagger}; it wins 97.8\,\% of validation games with the full history and is frozen. A second reader $R_2$ (Phi-4-mini, \citealp{abouelenin2025phi4mini}, same recipe; 77.8\,\% with the full history) tests whether trained states are informative to a reader they were not tuned to. The prompted formats are \fmt{summary}, \fmt{belief}, \fmt{slots}, \fmt{guide} (the summary prompt plus a hand-written rule: always keep food, tool and appliance locations, open containers and doors, and the room graph) and \fmt{lastk} (the most recent raw steps that fit), as in \S\ref{sec:related}; \fmt{oracle-b} is a privileged state recomputed from the game's true facts and cut to $B$ tokens, which shows what the budget can carry; \fmt{full} is the unbounded history. The prompts are reproduced in Appendix~\ref{app:formats}; \dssr uses the \fmt{summary} prompt.

\textbf{Protocol, statistics, pre-registration:}
Every cell is 45 (validation) or 90 (test) deterministic episodes per lag $\Fp \in \{0,10,20,40\}$; $\kappa$ is audited on up to six decision points per episode and 
every $\kappa$ reported per (format, lag) cell is the mean of per-game means.
Confidence intervals are paired percentile bootstraps over games (10{,}000 resamples), and all numbers in a table come from a single GPU.
Hypotheses, gates and the analysis plan were written before the corresponding data existed; two amendments made after the diagnostic (the recursive score and protocol $\Fp$) were recorded before the writer was trained, and the test split was opened once, after selection and seeds were fixed on validation (Appendix~\ref{app:gates}).

\subsection{The bottleneck is write-time regret, and it grows with lag}
\label{sec:diag}

\begin{table}[t]
\caption{Validation games under protocol $\Fp$, $B=128$: online success (\% of 45 games won) and mean write-time regret $\kappa$ (nats) per lag; success at $B=256$ for comparison. FULL has no budget; ORACLE-B is privileged ($\approx 47$ tokens). $\kappa$ is the mean of per-game means; $\beta$ lies between $-0.14$ and $0.23$ in every $B{=}128$ cell. Full grid with $\delta$, $\beta$, $\kappa$ and intervals in Appendix~\ref{app:stage2}.}
\label{tab:stage2}
\begin{center}
\small
\setlength{\tabcolsep}{4pt}
\begin{tabular}{l rrrr rrrr rrrr}
\toprule
& \multicolumn{4}{c}{success, $B{=}128$} & \multicolumn{4}{c}{$\kappa$, $B{=}128$} & \multicolumn{4}{c}{success, $B{=}256$} \\
\cmidrule(lr){2-5}\cmidrule(lr){6-9}\cmidrule(lr){10-13}
$\Fp$ & 0 & 10 & 20 & 40 & 0 & 10 & 20 & 40 & 0 & 10 & 20 & 40 \\
\midrule
\fmt{full} & 97.8 & 97.8 & 97.8 & 97.8 & -- & -- & -- & -- & -- & -- & -- & -- \\
\fmt{oracle-b} & 100 & 100 & 100 & 97.8 & $-0.19$ & $-0.17$ & $-0.21$ & $-0.15$ & 100 & 100 & 100 & 97.8 \\
\midrule
\fmt{lastk} & 0 & 0 & 0 & 0 & 0.71 & 1.80 & 1.46 & 1.46 & 2.2 & 0 & 0 & 0 \\
\fmt{summary} & 2.2 & 2.2 & 0 & 2.2 & 0.85 & 1.26 & 1.47 & 1.10 & 20.0 & 6.7 & 2.2 & 2.2 \\
\fmt{slots} & 11.1 & 0 & 0 & 0 & 0.49 & 1.55 & 1.38 & 1.69 & 15.6 & 4.4 & 2.2 & 0 \\
\fmt{belief} & 8.9 & 6.7 & 4.4 & 6.7 & 0.79 & 1.44 & 1.27 & 1.17 & 28.9 & 15.6 & 8.9 & 6.7 \\
\fmt{guide} & 11.1 & 11.1 & 2.2 & 6.7 & 0.66 & 0.72 & 0.84 & 1.10 & 40.0 & 31.1 & 31.1 & 22.2 \\
\bottomrule
\end{tabular}
\end{center}
\end{table}

\textbf{The budget suffices; the writer does not choose well:}
Table~\ref{tab:stage2} gives the validation grid (drawn in Appendix~\ref{app:stage2}, Figure~\ref{fig:f02}). With the full history the reader wins 97.8\,\% of the games at every lag; the privileged \fmt{oracle-b} state, about 47 tokens, wins 98--100\,\% at every budget. Every LLM-written state at $B=128$ wins 0--11\,\%, at $B=64$ none of them wins anything, and at $B=256$ the best, \fmt{guide}, falls from 40\,\% to 22\,\% as $\Fp$ grows. Decomposing the gap on 200--270 decision points per cell, $\kappa$ is 0.4--2.0 nats in every budgeted LLM cell (intervals exclude zero) while $|\beta| \le 0.40$; \fmt{oracle-b} has $\kappa<0$: a state written with knowledge of what matters beats the best of eight hindsight compressions of the whole history. Across the 76 (format, budget, lag) cells the mean gap $\delta$ predicts online success with Spearman $\rho=-0.84$ ($-0.90$ on the replicate GPU), so the offline loss is a valid proxy for the online outcome.

\textbf{Regret grows with rewrites since the relevant observation:}
Under protocol $\Fp$ the measured relevance lag $\ell$ rises from 4 to 16, 30 and 48 steps at $\Fp=0/10/20/40$, and $\kappa$ at $B=128$ rises from 0.5--0.9 nats to 1.1--1.8 for the four write-time formats (\fmt{guide}: $0.66 \to 1.10$): a large step within the first ten rewrites and a plateau after it, with $\beta$ flat. Against the measured lag, $\kappa$ rises from 0.61 \ci{0.38}{0.84} at $\ell \in [1,4]$ to 1.87--2.30 at $\ell \ge 10$ (Table~\ref{tab:ell}), while the full-history loss stays $\le 0.7$ and $\beta\approx 0$ in every bin, so the rise is writer loss, not task difficulty; $\kappa(\ell \ge 5) - \kappa(\ell<5) = +1.08$ \ci{0.24}{1.97}.

\textbf{What counts as lag:}
Table~\ref{tab:pilot} in Appendix~\ref{app:stage2} isolates the protocol result. The 40-step walk placed before the reveal leaves $\kappa$ unchanged ($-0.20$, $-0.34$ and $+0.24$ nats for three formats). Suppressing the re-printed room contents during that walk raises $\kappa$ at $F=40$ by $+0.47$ \ci{0.10}{0.93}: the walk had been refreshing the writer. Placing the walk after the reveal raises $\kappa$ by $+1.07$ \ci{0.42}{1.75}, while \fmt{full} and \fmt{oracle-b} stay at 98--100\,\%. Distractor steps are lag for a recursive writer only when they neither re-expose the relevant facts nor precede their revelation.

\begin{figure}[t]
\begin{center}
\includegraphics[width=0.42\linewidth]{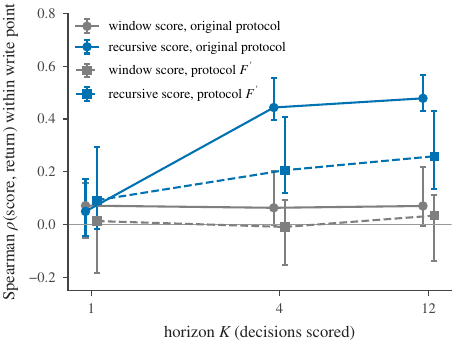}
\end{center}
\vspace{-4pt}
\caption{The sufficiency score predicts return only in its recursive form: mean within-write-point Spearman correlation between a candidate state's score $\SK$ and the realised return of the episode continued from it (200 write points $\times$ 8 candidates from \fmt{summary}/$B{=}128$ trajectories, continued to the end of the game; game-clustered 95\,\% intervals), for the window and the recursive score at $K \in \{1,4,12\}$ under both protocols.}
\label{fig:h2}
\end{figure}

\textbf{The sufficiency score predicts return only in its recursive form:}
We continued 200 write points $\times$ 8 candidate states from \fmt{summary}/$B{=}128$ trajectories to the end of the game (1{,}600 episodes) and correlated each score with the realised return within a write point (Figure~\ref{fig:h2}). The window score is uncorrelated with return at every horizon ($\rho \le 0.07$, both protocols) and uncorrelated with the recursive score ($\rho \le 0.09$). The recursive score reaches $\rho = 0.48$ \ci{0.39}{0.56} at $K=12$ ($0.44$ at $K=4$, $0.05$ at $K=1$) and $0.26$ \ci{0.13}{0.43} under $\Fp$, where 91 of the 200 write points have identical returns for all candidates. Continuing from the logged state reproduces the logged episode exactly (200/200), so the machinery is sound: the fixed-state approximation is invalid for a lossy writer because a state's value depends on how it will be rewritten.

\subsection{Training and selection}
\label{sec:train}

\textbf{Distillation:}
Each round collects 480 episodes on the training games (writer at temperature 0.7, reader $\epsilon$-greedy, $\Fp$ and $B$ drawn uniformly), samples 3{,}984 write points (half from Phase A), writes six candidates per point and scores them recursively. Supervised fine-tuning on the best candidate lifts validation success at $\Fp=0$ from 2.2\,\% to 13.3\,\%, leaves $\Fp \ge 10$ unchanged, and leaves $\kappa$ unchanged at every lag (mean over $\Fp$ 1.17 before, 1.20 after). The horizon matters already here: distilling the best candidate by $S_1$, $S_4$ or $S_{12}$ gives 4.4, 6.7 and 13.3\,\%. \fmt{outcome}'s round-0 target is degenerate: for 76\,\% of its write points all six candidates have the same return.

\textbf{Preference rounds and selection:}
Round 1 (4{,}460 pairs) takes $\kappa$ at $\Fp=0$ from 0.85 to 0.52 nats and raises it at $\Fp \ge 10$ (1.26/1.47/1.10 $\to$ 1.60/1.59/1.87); every variant behaves the same way, for a reason \S\ref{sec:credit} isolates on the training candidates. Policy iteration helps slowly: round 3 of the main line and of $K{=}4$ flatten the $\kappa$-versus-$\Fp$ curve (lag $\Fp \in \{20,40\}$: $-16$\,\% \ci{-34}{+7} and $-21$\,\% \ci{-40}{+2}), and no variant reduces lag $\kappa$ significantly on 45 games (Appendix~\ref{app:stage3}). The A4 variant, which samples candidates only where retention is still a choice (\S\ref{sec:method}), is the strongest: after two rounds it wins 26.7 / 11.1 / 4.4 / 2.2\,\% at $\Fp = 0/10/20/40$ (untrained 2.2 / 2.2 / 0 / 2.2), and its pair signal does not thin as the policy sharpens (80\,\% of write points with a qualifying pair, median gap 0.88 nats, against 54\,\% and 0.36 for the main line at round 3). Selection on validation success picks the A4 round-2 writer (11.1\,\% averaged over $\Fp$) ahead of $K{=}1$ round 1 (7.2\,\%), $K{=}4$ round 3 (6.7\,\%) and the main line's round 1 (5.0\,\%); its two additional seeds win 8.9 and 22.2\,\% at $\Fp=0$ and reduce zero-lag $\kappa$ by 28--64\,\% (Appendix~\ref{app:stage3}).

\subsection{Main comparison on the test split:}
\label{sec:main}
\begin{table}[t]
\caption{Test split (90 games), $B=128$, reader $R_1$: success (\%) and $\kappa$ (nats, mean of per-game means) per lag $\Fp$. Trained writers are the best validation round of each variant; \dssr is the selected A4 round-2 configuration, mean of three seeds (seeds at $\Fp=0$: 3.3 / 15.6 / 12.2). Intervals, $R_2$ rows and the budget frontier: Appendix~\ref{app:stage4}.}
\label{tab:test}
\begin{center}
\footnotesize
\renewcommand{\arraystretch}{0.95}
\begin{tabular}{l rrrr rrrr}
\toprule
& \multicolumn{4}{c}{success (\%)} & \multicolumn{4}{c}{$\kappa$ (nats)} \\
\cmidrule(lr){2-5}\cmidrule(lr){6-9}
$\Fp$ & 0 & 10 & 20 & 40 & 0 & 10 & 20 & 40 \\
\midrule
\fmt{full} (no budget) & 95.6 & 96.7 & 96.7 & 94.4 & -- & -- & -- & -- \\
\fmt{oracle-b} (privileged) & 98.9 & 100 & 98.9 & 97.8 & -- & -- & -- & -- \\
\midrule
\fmt{lastk} & 0 & 0 & 0 & 0 & -- & -- & -- & -- \\
\fmt{summary} & 3.3 & 2.2 & 1.1 & 0 & 0.79 & 1.05 & 1.35 & 1.66 \\
\fmt{belief} & 11.1 & 0 & 0 & 0 & 0.57 & 1.21 & 1.55 & 1.73 \\
\fmt{slots} & 10.0 & 0 & 0 & 0 & 0.85 & 1.75 & 1.64 & 1.96 \\
\fmt{guide} (hand-written rule) & \textbf{14.4} & \textbf{6.7} & 0 & 2.2 & \textbf{0.46} & \textbf{0.86} & 1.03 & \textbf{1.18} \\
\midrule
\dssr (A4 r2, 3 seeds) & 10.4 & 3.7 & 2.2 & 1.9 & 0.62 & 1.20 & 1.36 & 1.58 \\
\dssr main line (r1) & 7.8 & 1.1 & 1.1 & 1.1 & 0.51 & 1.15 & 1.44 & 1.92 \\
\fmt{outcome} (r3) & 8.9 & 0 & 1.1 & 0 & 0.57 & 1.39 & 1.52 & 1.84 \\
$K{=}1$ (r1) & 6.7 & 0 & 1.1 & 0 & 0.87 & 1.23 & 1.36 & 1.84 \\
$K{=}4$ (r3) & 11.1 & 1.1 & 2.2 & 1.1 & 0.54 & 1.08 & \textbf{0.96} & 1.47 \\
\fmt{no-hint} (r3) & 8.9 & 0 & \textbf{3.3} & \textbf{2.2} & 0.87 & 1.22 & 1.21 & 1.40 \\
\bottomrule
\end{tabular}
\end{center}
\end{table}

\textbf{Result:}
Table~\ref{tab:test} is the single-use evaluation. At zero lag the trained writer wins 10.4\,\% of the test games against 3.3\,\% for the prompted \fmt{summary} writer, $+7.0$ \ci{+1.9}{+12.2} points; every trained variant lands in the same band (6.7--11.1\,\%) and 
two of the three seeds are above the baseline (15.6 and 12.2 \%) and the selected seed ties it (3.3 \%). The comparison isolates training, since DSSR and SUMMARY differ only in the adapter; the BELIEF and SLOTS formats reach the same band at zero lag without training (11.1 and 10.0 \%).
Under $R_2$ the zero-lag ordering is reproduced (\fmt{guide} 11.1 $>$ \dssr 5.6 $>$ \fmt{summary} 2.2\,\%), so the trained states are informative to a reader they were not tuned to. Under lag the trained writer is not distinguishable from the prompted writers: at $\Fp \in \{20, 40\}$ it is $+1.5$ \ci{-0.4}{+3.3} points over \fmt{summary}, $+1.5$ \ci{+0.2}{+3.0} over \fmt{outcome} and $+0.9$ \ci{-0.9}{+2.8} over \fmt{guide}, the best prompted state there; its lag $\kappa$ ($\Fp \in \{20,40\}$) is 1.47 against 1.50 untrained ($-2$\,\% \ci{-16}{+16}). Every pre-registered gate on the lag gain fails (Appendix~\ref{app:gates}). The untrained writer's $\kappa$ rises with $\Fp$ on test ($0.79 \to 1.66$), reproducing the validation phenomenon, and \fmt{guide} 
has the lowest $\kappa$ of the untrained formats at every lag and the highest success averaged over lags
(at $B=256$: 30.0 / 16.7 / 15.6 / 15.6\,\%; Appendix~\ref{app:stage4}).
\begin{figure}[t]
\begin{center}
\includegraphics[width=0.6\linewidth]{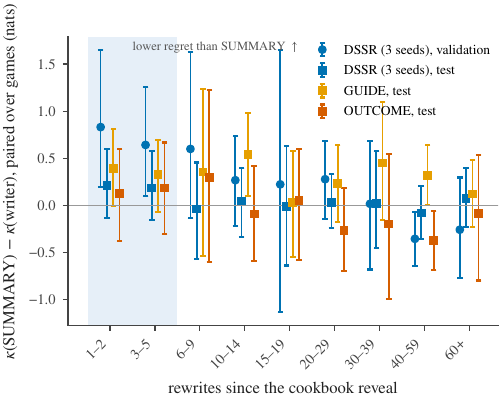}
\end{center}
\vspace{-4pt}
\caption{Where the reduction lives: paired difference of per-game mean $\kappa$ between the untrained \fmt{summary} writer and a trained or rule-based writer (positive = lower regret than \fmt{summary}), per bin of rewrites since the cookbook reveal, with 95\,\% game-bootstrap intervals; the three \dssr seeds are pooled per game. Shaded: bins where the trained writer's reduction is significant on validation (1--5 rewrites). Per-bin curves: Figures~\ref{fig:f08} and~\ref{fig:f09}.}
\label{fig:rewrites}
\end{figure}

\textbf{Where the reduction lives:}
Binning the audited decision points by the number of rewrites since the reveal (Figure~\ref{fig:rewrites}) gives the continuous form of the lag grid and the statement of the method's range. On validation the three seeds reduce $\kappa$ by 81\,\% within 1--2 rewrites (paired difference $+0.83$ nats \ci{+0.20}{+1.65}) and by 52\,\% within 3--5 (\ci{+0.10}{+1.25}), by 44\,\% at 6--9 (not significant), by about 20\,\% at 10--29, and not at all beyond, becoming \emph{worse} than the untrained writer at 40--59 rewrites; on test the profile is directional only ($-31$\,\% at 1--2, $-19$\,\% at 3--5, zero from 6 on). \fmt{guide} reduces $\kappa$ by 20--50\,\% in every bin up to 40--59 rewrites on test. The learned writer thus helps at short (1--2 rewrites) and medium (3--5) lag, weakly at 6--9, and not at 10 or more; a rule that names what to keep helps everywhere. An exploratory validation grid with eleven filler lengths agrees (Appendix~\ref{app:laggrid}; post hoc): the seeds' game-paired success gain over \fmt{summary} is $+17$ points at $\Fp=0$, $+7$ to $+10$ at $\Fp = 2$--$9$ (intervals above zero), $+5$ to $+7$ at $\Fp = 15$--$20$ and within noise at $\Fp = 30$--$40$.

\vspace{-4pt}
\subsection{Ablations}
\label{sec:ablations}
\vspace{-4pt}

The variants in Table~\ref{tab:test} were trained from the same candidates with the same recipe (intervals in Appendix~\ref{app:stage4}). Measured sufficiency beats the outcome reward on the same write problems by a small but significant margin under lag ($+1.5$ \ci{+0.2}{+3.0} points), and \fmt{outcome}'s $\kappa$ on test is \emph{higher} than the untrained writer's beyond ten rewrites. 
$K{=}4$ has the lowest lag $\kappa$ of the trained writers and $K{=}1$, the main line and \fmt{outcomes} the highest, but no pairwise difference is significant on 90 games.
\fmt{no-hint} keeps the zero-lag gain with a command-phrase fraction of 0.13--0.18 against 0.30--0.37, so the gain is not a memorised action list. At $B=64$ no LLM writer wins more than 2.2\,\%; at $B=256$ the trained writers' zero-lag gain widens but \fmt{guide} still dominates. Round 0 supplies most of the zero-lag gain, and the test seed spread at zero lag (3.3--15.6\,\%) is as large as the gain.

\vspace{-4pt}
\subsection{Why the gain shrinks with delay: credit assignment across rewrites}
\label{sec:credit}
\vspace{-4pt}

\begin{figure}[t]
\begin{center}
\includegraphics[width=0.44\linewidth]{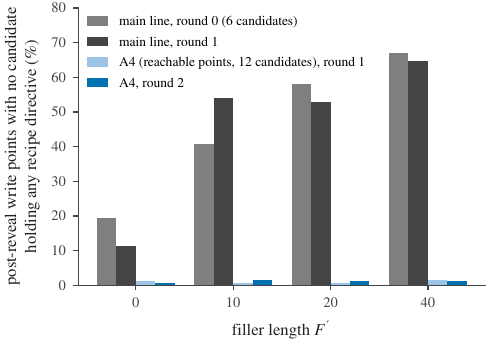}
\end{center}
\vspace{-4pt}
\caption{Credit assignment under lag: fraction of post-reveal training write points at which no candidate state holds any recipe directive, per $\Fp$ (main line, rounds 0--1; A4, rounds 1--2). Once a rewrite has dropped the recipe, best-of-$n$ has nothing to select; A4's reachability filter removes the case by construction. Coverage of the best and worst candidate: Figure~\ref{fig:f07}.}
\label{fig:credit}
\end{figure}

The reduction of regret is largest within a few rewrites of the reveal and shrinks after (Figure~\ref{fig:rewrites}).
Two measurements on the round-0 training candidates explain this profile (2{,}818 post-reveal write points; coverage is the fraction of the recipe's verb--ingredient directives present in a state; Figure~\ref{fig:credit} and Appendix~\ref{app:stage3}).
First, candidates inherit their input.
The six candidates are rewrites of the same previous state, and their mean coverage (0.24) equals its coverage (0.24); at 51\,\% of post-reveal write points (67\,\% at $\Fp=40$) no candidate contains any directive, so once a rewrite has dropped the recipe, best-of-$n$ cannot select what is absent.
Second, the score is taken under a forgetful continuation.
Where candidates do differ in coverage (9\,\% of points), $S_{12}$ prefers the more complete one only at chance level (28\,\% against 29\,\%): the candidate is handed to the current writer, which drops the recipe within a few rewrites, while the decisions that would reward keeping it are 10--40 steps away.
Both effects grow with delay, and the training signal thins accordingly: the median best-minus-worst score gap is 0.6--0.8 nats at Phase-A write points but 0.23--0.27 nats at $\Fp \ge 10$.
A per-step score therefore gives little credit for retention under a lossy recursive policy: keeping a fact at step $t$ is worth something only if it is also kept at $t{+}1, \dots, t{+}40$, and the current writer does not keep it.
More data of the same kind does not change either effect.

The two remedies we tried confirm the diagnosis without removing it.
Policy iteration makes retention slightly more visible: in the round-1 candidates the score prefers the more complete candidate 35\,\% against 26\,\%.
Reachability-restricted sampling (A4) removes the inherited-input case by construction (1\,\%) and restores a strong preference signal, which is why it is the selected line, but its lag $\kappa$ stays at the untrained level on validation ($+3$\,\% \ci{-18}{+29} at $\Fp \in \{20,40\}$ for the selected seed, $-3$\,\% \ci{-19}{+20} for the three seeds pooled) and on test.
A hand-written retention rule does not face this problem: it is applied at every rewrite, so it keeps a fact for as long as the rule names it, and its reduction of regret does not decay (20--50\,\% in every rewrite bin on test; Figure~\ref{fig:rewrites}).
What is missing is an objective that credits keeping a fact at $t$ with its value beyond the current writer's forgetting horizon (\S\ref{sec:conclusion}).

\vspace{-4pt}
\section{Limitations}
\label{sec:limitations}
\vspace{-4pt}
Our environment is a controlled diagnostic: Phase A is scripted, and under protocol $\Fp$ the writer knows the recipe matters, so $\Fp$ tests retention of a known-relevant fact rather than the prospective choice of what will matter.
\fmt{guide}, which uses task knowledge, remains the strongest budgeted state, and \dssr helps only at short and medium delay.
$\kappa$ is relative to a best-of-eight hindsight compression and a behaviour-cloned reader, and the second reader misses its pre-registered bar by one game.
We study only 4B-parameter models, long delays on Phase-A facts are rare, validation selection picked the weakest test seed, and replicates on a second GPU differ by up to 14 points, so each table uses one GPU.
\vspace{-4pt}
\section{Conclusion and future work}
\label{sec:conclusion}
\vspace{-4pt}
With a fixed budget, a constant-context agent loses mainly through write-time regret, not the budget, and the regret grows with the number of rewrites a fact must survive.
A candidate state is worth what the reader can do from it after the writer has rewritten it: this recursive score predicts return where the fixed-state approximation does not, and a writer trained on it keeps recently observed facts, with a significant gain on held-out test games when facts are needed soon.
The gain shrinks with delay, because a per-step objective cannot credit keeping a fact whose payoff lies beyond the current writer's forgetting horizon; a hand-written rule that uses task knowledge still does better there.
Future objectives should score a candidate under a continuation that keeps what the candidate keeps, for example by enforcing retention in the roll-forward, training writer and roll-forward jointly, or following a curriculum from short to long delays; readers that can request facts back are the alternative on the reader side.
The decomposition, the protocol and the pre-registered evaluation give a way to measure these proposals.

\newpage
\bibliography{iclr2027_conference}
\bibliographystyle{iclr2027_conference}

\clearpage
\begin{appendix}
\appendix
\input{appendix}
\end{appendix}
\end{document}

%% file: math_commands.tex
\usepackage{amsmath,amsfonts,bm}

\def\eqref#1{equation~\ref{#1}}

\def\1{\bm{1}}

\DeclareMathAlphabet{\mathsfit}{\encodingdefault}{\sfdefault}{m}{sl}
\SetMathAlphabet{\mathsfit}{bold}{\encodingdefault}{\sfdefault}{bx}{n}



%% file: 0_abs.tex
Long tasks produce more history than an LLM agent can hold in its context, and more than it uses reliably even when the history fits.
A growing line of work therefore has agents carry a short written state instead: at every step a writer rewrites the state, and a reader acts from the state alone.
Steps stay cheap, but anything the writer drops is lost before later decisions reveal that they need it. 
We quantify this loss and ask whether training can reduce it.
Comparing the written state with the best state of the same size written in hindsight, we split the reader's loss into a budget loss, which any state of that size must incur, and a write-time regret, which comes from the writer's choices.
In TextWorld cooking games where we control how long a fact must be carried before it is needed, a 128-token state holding the facts wins nearly every game, 
while prompted language-model writers win at most 17\,\%.
Almost all of the loss is write-time regret, and it grows with the delay.
Creating a real delay takes care: steps inserted between a fact and its use delay it only if they do not show the fact again.
We then train the writer from the reader's own loss.
DSSR (decision-sufficient state representations) scores candidate states by how well the reader acts after the writer carries them forward, and teaches the writer to prefer the better ones.
This forward-rolled score predicts game outcomes ($\rho = 0.48$; $0.26$ under the delayed protocol), whereas scoring a candidate as a fixed context, as hindsight methods usually do, does not ($\rho \le 0.07$).
On a pre-registered test split opened once, training adds $+7.0$ $[+1.9, +12.2]$ points of success when facts are needed soon, 
bringing a plain summary writer to the level of belief- and slot-based memory prompts.
The gain shrinks as the delay grows: on held-out games it is only at the shortest delay, and our pre-registered tests for a gain under long delay fail. A hand-written rule that names what to keep, using task knowledge the writer is not given, does better at every delay.
We trace this limit to credit assignment: keeping a fact now pays off only if every later rewrite keeps it too, which a per-step score cannot see.
Anonymous codebase: \url{https://anonymous.4open.science/r/AgentDSSR}.

%% file: 1_intro.tex
\section{Introduction}
\label{sec:intro}

\begin{figure}[t]
\begin{center}
\resizebox{0.9\linewidth}{!}{\input{figures/fig1_teaser}}
\end{center}
\caption{Write-time compression under relevance lag. (a)~One episode of a constant-context agent: at every step the writer $U$ rewrites the state $z_t$ ($\le B$ tokens) from $z_{t-1}$ and the new observation $o_t$, and the history $h_t$ is discarded. A fact observed at step 3 (the knife's location) is needed at step 7 (the recipe says \emph{slice}), $\ell=4$ rewrites later; whether $z_6$ still holds it was decided by $U$ before the need was known. (b)~Read-time compression (prompt compression, retrieval-based memory) keeps the history and compresses once the query is known; write-time compression (this work) compresses before the decision is known, recursively and under budget: what $z_t$ drops is gone for every later decision.}
\label{fig:teaser}
\end{figure}

LLM agents outgrow their context windows: a household task, a web session or a text game can run for hundreds of steps and many thousands of tokens of observations.
A widely used remedy bounds the agent's working memory \citep{wang2023recursive,packer2023memgpt,zhou2025mem1,singh2026agentbrace}.
A writer condenses the previous memory and each new observation into a fixed-size text state, and a reader selects actions from that state and the current observation, so every step costs the same however long the episode has run.
The price is irreversibility.
Raw observations are dropped after each rewrite, and a fact reaches the step that needs it only if every intermediate rewrite has preserved it (Figure~\ref{fig:teaser}a).
Prompt compression and retrieval-based memory avoid this by keeping the raw record and compressing it once the query is known (Figure~\ref{fig:teaser}b).
A constant-context agent must commit at write time; this paper measures what that commitment costs and asks whether the writer can be trained to make it well.

Two things can go wrong: the state can be too small to hold the facts that matter, or the writer can fill it with the wrong facts.
Benchmarks that report only task success \citep{maharana2024locomo,wu2025longmemeval,hu2026memoryagentbench} cannot tell these apart, yet only the second can be fixed by a better writer.
We separate them (\S\ref{sec:setting}) by comparing the reader's decisions under three inputs: the full history, the best state of the same size that could have been written in hindsight, and the state the writer actually wrote.
The first difference is the \emph{budget loss}, what any state of that size must give up; the second is the \emph{write-time regret}, what the writer gave up by choosing without knowing the future.

We measure both in TextWorld cooking games \citep{cote2018textworld,adhikari2020gata}, where we control how long a fact must be carried before it is needed.
The budget is not the problem: a 128-token state holding the true facts wins nearly every game, while every state written by a language model under five prompt designs modelled on published memory systems wins at most 17\,\%.
Almost all of the loss is write-time regret, and it grows the longer a fact must survive repeated rewriting.
Controlling this delay turned out to be subtle: steps inserted between a fact and its use create a real delay only if they do not show the fact again (\S\ref{sec:diag}).

We then train the writer with \dssr (decision-sufficient state representations; \S\ref{sec:method}).
At a write step the writer proposes several candidate states, the frozen reader scores each by how well it makes the next few decisions from it, and the writer learns to prefer the better candidates.
What matters is how a candidate is scored: the reader must see the state as the writer will actually carry it forward, not the candidate followed by the raw steps, as hindsight methods usually assume.
Only the first form predicts how the game ends ($\rho = 0.48$ against $0.07$), because a state's value depends on what the writer will later drop from it.

On a test split opened once, after every choice was fixed, the trained writer wins $+7.0$ \ci{+1.9}{+12.2} points more games than the same writer without training when facts are needed soon after they are seen, to the level of the belief- and slot-based prompts (\S\ref{sec:main}).
The gain shrinks as the delay grows and is significant on held-out games only when facts are needed soon; our pre-registered tests for a gain under long delay fail.
A hand-written rule that tells the writer which kinds of facts to keep helps at every delay, but it relies on knowledge of the task that the trained writer does not receive.
We trace the limit to credit assignment (\S\ref{sec:credit}): keeping a fact now pays off only if every later rewrite keeps it too, and a per-step score cannot see that.

\textbf{Contributions:}
(i)~A decomposition of a constant-context agent's decision loss into budget loss and \emph{write-time regret}, a protocol verified to create delay, and the finding that the writer's choices, not the budget, are the bottleneck.
(ii)~A \emph{recursive sufficiency score} that evaluates a candidate state by rolling the writer forward through it, and evidence that only this form predicts outcomes.
(iii)~\dssr, which trains the writer from this score, with a significant gain on a single-use test split when facts are needed soon and a gain that shrinks with delay.
(iv)~A diagnosis of why a per-step objective cannot teach long-range retention.

%% file: figures/fig1_teaser.tex
\begin{tikzpicture}[
  font=\footnotesize, >=Latex, line cap=round,
  obs/.style={draw=black!70, rounded corners=1.5pt, fill=white, minimum width=8mm, minimum height=5mm, inner sep=1.5pt},
  gone/.style={draw=cGray!70, rounded corners=1.5pt, fill=black!4, text=cGray, minimum width=8mm, minimum height=5mm, inner sep=1.5pt},
  state/.style={draw=cStateLine, thick, fill=cState, minimum width=9mm, minimum height=6mm, inner sep=1pt},
  writer/.style={draw=cWriterLine, fill=cWriter, rounded corners=2pt, minimum width=3.6mm, minimum height=3.6mm, inner sep=0.5pt, font=\scriptsize\bfseries},
  fact/.style={draw=cFact, thick, fill=cFactFill},
  dec/.style={draw=cDec, thick, fill=cDecFill},
  lab/.style={font=\scriptsize},
  small/.style={font=\tiny},
]
\def\dx{1.85}
\def\yo{1.45}
\def\ys{0}
\def\wx{0.83}
\node[state] (s0) at (0,\ys) {$z_0$};
\foreach \i [evaluate=\i as \j using int(\i-1)] in {1,...,7} {
  \node[writer] (w\i) at ({\dx*\i-\wx},\ys) {$U$};
  \node[state]  (s\i) at ({\dx*\i},\ys) {$z_{\i}$};
  \draw[->, thick, cStateLine] (s\j) -- (w\i);
  \draw[->, thick, cStateLine] (w\i) -- (s\i);
}
\foreach \i in {1,2,4,5,6} {
  \node[gone] (o\i) at ({\dx*\i-\wx},\yo) {$o_{\i}$};
  \draw[->, black!50] (o\i.south) -- (w\i.north);
}
\node[obs,fact] (o3) at ({\dx*3-\wx},\yo) {\textcolor{cFact}{$\boldsymbol{o_3}$}};
\node[obs,dec]  (o7) at ({\dx*7-\wx},\yo) {\textcolor{cDec}{$\boldsymbol{o_7}$}};
\draw[->, cFact, very thick] (o3.south) -- (w3.north);
\draw[->, cDec,  thick] (o7.south) -- (w7.north);
\node[small, text=cFact, align=center, above=1mm of o3] {``a \textbf{knife} is in the drawer''};
\node[small, text=cDec, align=center, above=1mm of o7] {recipe says \textbf{slice} the carrot\\ $\Rightarrow$ agent must fetch the knife};
\begin{scope}[on background layer]
  \node[fit=(o1)(o6), draw=none, fill=black!3, rounded corners=2pt, inner sep=2.5pt] (hist) {};
\end{scope}
\node[small, text=cGray, anchor=south west] at ($(hist.north west)+(0,0.02)$) {history $h_t$ is \textbf{discarded}};
\node[small, text=cDec, align=center, below=0.6mm of s6] {does $z_6$ still\\ hold ``knife''?};
\draw[decorate, decoration={brace, mirror, amplitude=2pt}, cStateLine]
  ($(s3.south west)+(0,-0.7mm)$) -- ($(s3.south east)+(0,-0.7mm)$) node[midway, below=2pt, lab, text=cStateLine] {$|z_t|\le B$};
\draw[decorate, decoration={brace, amplitude=3pt}, thick]
  ($(o3.north)+(0,0.85)$) -- ($(o7.north)+(0,0.85)$) node[midway, above=3pt, lab] {\textbf{relevance lag} $\ell=4$};
\draw[->, black!70] (-0.4,-0.95) -- ({\dx*7+0.7},-0.95) node[right, lab] {$t$};
\foreach \i in {1,...,7} { \draw[black!70] ({\dx*\i},-0.9) -- ({\dx*\i},-1.0); }
\node[lab, below, text=cFact] at ({\dx*3},-1.0) {write $z_3$};
\node[lab, below, text=cDec]  at ({\dx*7},-1.0) {decide};
\node[anchor=west, font=\small\bfseries] at (-0.55,{\yo+1.25}) {(a)};
\begin{scope}[shift={(0,-2.2)}]
  \node[anchor=west, font=\small\bfseries] at (-0.55,0.05) {(b)};
  \begin{scope}[shift={(0.1,0)}]
    \node[lab, anchor=west] at (0,0) {\textbf{Read-time} compression (prompt compression, retrieval-based memory)};
    \node[draw=black!70, fill=black!4, minimum width=27mm, minimum height=5.5mm, anchor=west, font=\scriptsize] (rh) at (0,-0.6) {full history $h_t$ kept};
    \node[writer, anchor=west, minimum width=4mm, minimum height=4mm] (rc) at (3.4,-0.6) {C};
    \node[state, anchor=west, minimum width=8mm] (rz) at (4.1,-0.6) {\scriptsize ctx};
    \node[obs, dec, anchor=north, font=\scriptsize] (rq) at ($(rc.south)+(0,-0.3)$) {query $q$};
    \draw[->, thick] (rh) -- (rc);
    \draw[->, cDec, thick] (rq.north) -- (rc.south);
    \draw[->, thick, cStateLine] (rc) -- (rz);
    \node[small, anchor=north west, text=black!70, text width=6.4cm, align=left] at (0,-1.55) {compresses \emph{after} the query is known; nothing is lost first.};
  \end{scope}
  \begin{scope}[shift={(7.35,0)}]
    \node[lab, anchor=west] at (0,0) {\textbf{Write-time} compression (this work)};
    \node[state, anchor=west, minimum width=8mm] (ws0) at (0,-0.6) {$z_{t-1}$};
    \node[obs, anchor=west, minimum width=6mm] (wo) at (1.05,-0.6) {$o_t$};
    \node[writer, anchor=west, minimum width=4mm, minimum height=4mm] (ww) at (1.95,-0.6) {$U$};
    \node[state, anchor=west, minimum width=8mm] (ws1) at (2.6,-0.6) {$z_t$};
    \node[obs, dec, anchor=west, font=\scriptsize, dashed] (wq) at (4.75,-0.6) {decision at $t{+}\ell$};
    \draw[->, thick, cStateLine] (ws0) -- (ww);
    \draw[->, black!60] (wo) -- (ww);
    \draw[->, thick, cStateLine] (ww) -- (ws1);
    \draw[->, dashed, cDec] (ws1) -- node[small, above=-0.5pt, pos=0.5, text=cDec] {$+\ell$ steps} (wq);
    \draw[cGray, line width=0.8pt] ($(ws0.north west)+(-0.06,0.1)$) -- ($(wo.south east)+(0.06,-0.1)$);
    \node[small, text=cGray, anchor=north west] at (0,-0.95) {$h_t$ discarded once $z_t$ is written};
    \node[small, anchor=north west, text=black!70, text width=6.4cm, align=left] at (0,-1.3) {compresses \emph{before} the decision is known, recursively, under budget $B$: what $z_t$ drops is gone for all later decisions.};
  \end{scope}
\end{scope}
\end{tikzpicture}

%% file: appendix.tex

\section{Environment and protocol}
\label{app:env}

\paragraph{Games.}
TextWorld cooking games \citep{cote2018textworld} generated with \texttt{tw-cooking --recipe 3 --take 3 --go \{6,9,12\} --open --cook --cut} (no \texttt{--drop}), textworld 1.7.0. Counts per number of rooms: 80 / 15 / 30 for train / validation / test (240 / 45 / 90 games), disjoint seed ranges per split; the \texttt{--split} setting changes the food-item distribution, so validation and test games contain unseen foods and preparations. Every game has a maximum score of 11. Games are generated once with a fixed hash seed, archived with their md5 and the exact generation command, and never regenerated. The environment exposes, per step, the cleaned feedback text, the inventory sentence (appended to every observation, since \texttt{inventory} is not an allowed action and inventory tracking is a short-lag skill that would otherwise confound the long-lag phenomenon), the admissible commands, and the privileged fields (facts, entities, recipe, walkthrough and the optimal policy \texttt{policy\_commands}). The privileged fields define $a^*$ and \fmt{oracle-b} and are never shown to the writer or the reader.

\paragraph{Phase A.}
(1)~\emph{Explore}: a depth-first search over rooms that discovers the map from the room headers only; on the first visit of a room it executes every admissible \texttt{open} of a non-door container (so that contents are printed), then every admissible \texttt{open} of a door, then tries the exits in the fixed order north, south, east, west, backtracking every edge (the walk ends in the start room); it issues only \texttt{go} and \texttt{open} and never \texttt{examine}. (2)~\emph{Reveal}: the shortest route to the kitchen and \texttt{examine cookbook}, whose observation is the recipe (ingredients and directions). (3)~\emph{Filler} (protocol $\Fp$): a random walk of $\Fp \in \{0,10,20,40\}$ \texttt{go} steps among visited rooms, drawn from a generator seeded by the game id, so the walks for different $\Fp$ are nested prefixes of one walk and $\Fp$ changes only how much of it is inserted. The original pre-registered order placed the filler between exploration and the reveal (\S\ref{sec:setting}); Appendix~\ref{app:stage2} keeps those results. \emph{Validity check}: every recipe ingredient and every required tool or appliance (\texttt{slice/chop/dice} $\to$ knife, \texttt{fry} $\to$ stove, \texttt{roast} $\to$ oven, \texttt{grill} $\to$ BBQ) must occur in a Phase-A observation other than the cookbook; all 285 train and validation games pass, and the explorer reaches every room in every game.

\paragraph{Phase B.}
The reader acts until the game is won or lost, until the optimal policy is empty (the game has become unwinnable, which TextWorld reports after a wrong cut or a burnt ingredient), or after 50 actions. Options are the admissible commands minus \texttt{look}, \texttt{inventory}, \texttt{examine} (except \texttt{examine cookbook}), \texttt{close}, \texttt{put} and \texttt{insert}; $a^*$ is appended if missing (logged); the list is shuffled with a string seed of game and step and labelled \texttt{A--Z, a--z}. $a^*$ is the first command of \texttt{policy\_commands}, which re-plans after deviations. Success is winning; the normalised score is score / 11. Episodes are advanced in lock-step across games: one batched writer call, then one batched reader call, per global step. The writer writes at every step of both phases, starting from $z_0 = U(g, \text{(empty)}, \text{(none)}, o_0)$.

\paragraph{Environment statistics.}
Table~\ref{tab:envstats} gives Phase-A lengths and history sizes on the train and validation games (complete oracle episodes; the oracle wins every game). Phase-A length varies with the map even at fixed rooms and $\Fp$ because the explorer backtracks every edge and the reveal route depends on where the walk ends; the per-game length is logged with every episode.

\begin{table}[ht]
\caption{Environment statistics per (rooms, $\Fp$) on the 285 train and validation games: mean Phase-A length (steps), mean Phase-B length of the oracle, mean and maximum full-episode history length in tokens of the reader tokenizer.}
\label{tab:envstats}
\begin{center}
\small
\begin{tabular}{rr rr rr}
\toprule
rooms & $\Fp$ & Phase-A steps & Phase-B steps & history tokens (mean) & history tokens (max) \\
\midrule
6 & 0 & 15.7 & 12.0 & 2230 & 2821 \\
6 & 40 & 55.7 & 12.0 & 6843 & 9087 \\
9 & 0 & 25.0 & 18.9 & 4151 & 5514 \\
9 & 40 & 64.8 & 18.7 & 9102 & 10908 \\
12 & 0 & 33.5 & 21.1 & 4825 & 6721 \\
12 & 40 & 73.3 & 20.8 & 9079 & 11591 \\
\bottomrule
\end{tabular}
\end{center}
\end{table}

\section{Models, serving and readers}
\label{app:reader}

\paragraph{Serving.}
Writer and reader $R_1$ share the base model \texttt{Qwen/Qwen3-4B-Instruct-2507} \citep{yang2025qwen3} with separate LoRA adapters; $R_2$ is \texttt{microsoft/Phi-4-mini-instruct} \citep{abouelenin2025phi4mini}. Inference runs in vLLM 0.29 \citep{kwon2023vllm} with prefix caching and batch-invariant kernels (\texttt{VLLM\_BATCH\_INVARIANT=1}); without them we measured differences of up to 0.125 nats for the same prompt in different batches, with them the same item scored alone and inside a batch of unrelated items gives bit-identical log-probabilities (unit-tested). Token budgets and state lengths are measured with the $R_1$ tokenizer. Model revisions are pinned and recorded in every run.

\paragraph{Reader prompt and scoring.}
The reader prompt is: a generic system instruction (reproduced in Appendix~\ref{app:formats}), then \texttt{Goal:}, the \texttt{CONTEXT} block (full history, written state or \fmt{lastk} block), \texttt{Current observation:}, \texttt{Options:} (one lettered command per line) and \texttt{Action:}; the answer is the bare label as the first assistant token under the chat template. Every label is a single, prefix-stable token for all tokenizers used (52/52, re-verified per tokenizer). The action distribution is one forward pass with the logits restricted to the shown labels and renormalised, so $\NLL_t = -\log p_\pi(\text{label of } a^*_t)$ needs no further normalisation. Evaluation acts by argmax; data collection uses $\epsilon$-greedy with $\epsilon=0.1$ from a seeded generator. The \fmt{lastk} block holds the most recent raw past steps that fit in $B$ tokens followed by the last action, never exceeding $B$ (decode/re-encode guard).

\paragraph{Reader training.}
Behaviour cloning of $a^*$ on Phase-B decision points of the 240 training games. The context of each example is drawn from 19 variants: the full history (weight 25\,\%) and \fmt{lastk}, \fmt{summary}, \fmt{belief}, \fmt{slots}, \fmt{guide} and \fmt{oracle-b} at $B \in \{64,128,256\}$ (the remaining 75\,\% uniformly), one variant per episode, then a stratified sample of 6{,}000 decision points per DAgger iteration \citep{ross2011dagger} with exactly these weights and a prefix-preserving ordering. \fmt{oracle-b} is included so that the feasibility comparison (\fmt{oracle-b} versus \fmt{full}) is not confounded by an unfamiliar format; the trained \dssr writer is never part of the mixture, and the recursive-format states come from the untrained writer at temperature 0. Iteration 1 drives Phase B with $a^*$; iteration 2 drives it with the iteration-1 reader ($\epsilon=0.1$) and relabels every visited decision point with the $a^*$ of the state actually reached; the final reader is trained on the union. The loss is the cross-entropy of the label token at the final prompt position only (full vocabulary), which lets full-history contexts of up to 11.6k tokens be trained without truncation on a 32\,GB GPU (bf16 base, LoRA rank 32 on all linear layers, gradient checkpointing, effective batch 16, AdamW, learning rate $10^{-4}$ with 3\,\% warm-up and linear decay, 2 epochs). Held-out monitoring uses about 5\,\% of the training \emph{games}. Model size and the number of examples were chosen from a 2k/4k/6k learning curve on validation.

\paragraph{Acceptance.}
The pre-registered acceptance criterion is at least 80\,\% online success with the full history on the validation games at zero lag, averaged over the three room counts. $R_1$: 88.9\,\% after iteration 1 and 97.8\,\% after iteration 2 (100 / 100 / 93.3\,\% at 6 / 9 / 12 rooms; \texttt{results/stage1/success\_valid\_r1\_it2\_FULL\_F0}); top-1 accuracy on held-out validation decision points 0.989 with the full history, $\ge 0.984$ with the $\le 64$-token oracle state, and 0.54--0.86 with states from the untrained writer. $R_2$ (Phi-4-mini, same recipe): 68.9\,\% after iteration 1, 77.8\,\% after iteration 2 (93 / 87 / 53\,\%) and 77.8\,\% after a third iteration (100 / 87 / 47\,\%; accuracy 0.736 / 0.752 / 0.740), one game short of the bar in both; Phi-4-mini plateaus on the 12-room games. We use iteration 2 for the transfer test and state the shortfall (\S\ref{sec:limitations}). Generating $R_2$'s reader-driven data requires the Qwen writer and the Phi reader to be resident at once (Amendment A3, Appendix~\ref{app:gates}); the two engines each take half of the 96\,GB GPU.

\section{State formats and prompts}
\label{app:formats}

\paragraph{Writer prompt.}
Every recursive format is a chat prompt whose system message is the format's prompt file with the budget substituted, and whose user message is \texttt{Goal:} / \texttt{Previous state:} (\texttt{(empty)} at step 0) / \texttt{Last action:} (\texttt{(none)} at step 0) / \texttt{New observation:}. The writer request carries no field for the options, for $a^*$ or for any privileged information, and every writer message is checked structurally against the reader's layout before it is sent. Generation uses $\texttt{max\_tokens}=B$; the output is re-tokenised with the reader tokenizer and, if longer than $B$, cut to $B$ tokens and decoded, with the cut lowered by one token per round until the decoded text re-encodes to at most $B$ tokens. Every state stored or shown to the reader is therefore at most $B$ tokens by construction. The four prompted formats share the framing and differ only in the retention instruction. The \fmt{summary} prompt (also used by \dssr, whose only difference is the adapter) reads:

{\small
\begin{quote}
You are the memory writer for an agent acting in an interactive text environment. The agent cannot see its past observations: the state you write is the only record of everything that has happened so far, and it is the only memory the agent will have from now on. Anything you leave out is lost for good.

Update the summary of what has happened so far. Combine the previous state, the last action and the new observation into one updated summary. Keep whatever will matter later and drop what will not.

Rules:\\
- The state must fit in \{budget\} tokens. Use short, dense phrasing.\\
- Write only the state text. No preamble, no headings, no commentary, no questions.
\end{quote}
}

\fmt{guide} adds one paragraph after the second: ``Environment guideline: always keep the locations of food items, tools (knife) and appliances (stove, oven, BBQ), which containers/doors are open, and how rooms connect. Drop other details before dropping any of these.'' \fmt{belief} asks for one atomic claim per line, each ending with \texttt{[certain]}, \texttt{[likely]} or \texttt{[unsure]}, updated by adding, correcting or removing claims (Agent-BRACE-like, \citealp{singh2026agentbrace}). \fmt{slots} asks for exactly three labelled slots, \texttt{Progress:}, \texttt{Attempted actions:} and \texttt{Saved observations:} (PABU-like, \citealp{jiang2026pabu}).

\paragraph{Reader system prompt.}
{\small
\begin{quote}
You are an agent playing a text-based cooking game. You complete the goal by choosing one action at a time.

Each turn you receive:\\
- Goal: what you must accomplish.\\
- CONTEXT: what has happened so far. It may be the full history or a compressed state, and it may be incomplete.\\
- Current observation: what you see right now.\\
- Options: the admissible commands, each with a letter label.

Choose the single option that best makes progress toward the goal, using the context and the current observation. Answer with only the letter of the chosen option and nothing else.
\end{quote}
}
It contains no recipe knowledge and no retention advice, and the same prompt is used for every format and both readers.

\paragraph{\fmt{oracle-b}.}
The privileged state is recomputed at every step from the game's true facts and rendered to fit the budget in priority order: the current room; for every ingredient not yet carried and fully prepared, its location and the preparation verbs still to apply; the knife while a cut remains; an appliance while its cooking verb remains; the route from the current room to the room of the next reference step, then, if the budget allows, the routes to the other needed rooms. Lines are selected whole; when the long form does not fit, container names are dropped. After the meal is prepared the state is the single fact that the meal is carried and must be eaten. A unit test checks at every decision point of oracle trajectories, at every budget, that all core lines are present. Its mean length is about 47 tokens. \fmt{oracle-b} is never shown to the writer and is excluded from every writer-training set.

\paragraph{Read-time oracle $\zrt$.}
The untrained writer with the \fmt{summary} prompt, \texttt{Previous state: (empty)}, \texttt{Last action: (none)}, the entire rendered history $h_t$ as the observation block, $N=8$ samples at temperature 1.0 with the same budget enforcement; the sample with the lowest $\NLL_t$ is kept. Selection is the only place where $a^*$ enters the writer side, and it makes $\zrt$ a hindsight reference rather than a bound.

\section{Metrics, statistics and provenance}
\label{app:metrics}

\paragraph{Decomposition and lag.}
$\delta$, $\beta$ and $\kappa$ are computed element-wise from $\NLL_t(z^M_t)$, $\NLL_t(h_t)$ and $\NLL_t(\zrt_t)$ on up to six randomly chosen Phase-B decision points per episode (fixed seed); the identity $\delta=\beta+\kappa$ holds to rounding and is unit-tested. The relevance lag uses leave-one-chunk-out re-scoring of the full history (chunk = one room visit in Phase A, five-step blocks in Phase B) on 900 sampled decision points of the $B=128$ write-time formats; $\ell_t = t - s^*$ for the chunk $s^*$ whose removal increases $\NLL_t(h_t)$ most, undefined when the largest increase is below 0.5 nats (0.3 and 1.0 are reported as robustness checks). Ties go to the earliest chunk.

\paragraph{Bootstrap.}
Every interval is a paired percentile bootstrap over games: games are resampled with replacement (10{,}000 resamples, seed 0), the same resampled games are used in both conditions of a paired comparison, and the interval is the 2.5--97.5\,\% range of the resampled statistic; the point estimate is the plug-in statistic. Per-game values (mean $\kappa$ over the game's audited points, or 0/1 success) are computed first, so every game has the same weight, and the game sets of two compared conditions must be identical; every $\kappa$ reported per cell in the tables and figures is this mean of per-game means (the run summaries' pooled means over points differ from it by up to 0.26 nats on the test split). A lag value over $\Fp\in\{20,40\}$ is the mean of a game's two per-lag means (each lag weighs the same), so it equals the average of the two cells, and relative changes are ratios of game means recomputed in every resample. 

For the continuation study the write point is the resampling unit, with game-clustered resampling. The $\kappa$-versus-$\Fp$ slope is the pooled least-squares slope over all (game, $\Fp$) cells, with games resampled as units; the difference of two slopes uses the same resamples. 
Tables 7 and 15 average over the decision points of a bin; the paired contrasts of Table 17 and Figure 3 first average per game within a bin.
\paragraph{Privileged-information check.}
Every run stores the first 200 reader prompts and the first 200 writer prompts verbatim (drawn with a seeded rule that spreads them over episodes). A time-aware string search looks for every line of the recipe, the joined walkthrough and policy command lists, every privileged fact string and the \fmt{oracle-b} lines of every visited state; a string is exempt for a prompt at step $t$ only if it occurs verbatim in an observation of a step $\le t$, so recipe lines count as leaks before the cookbook is read and as legitimate afterwards. Positive controls are part of the test suite. The only exempt prompts are \fmt{oracle-b}'s reader prompts, which are privileged by design.

\paragraph{Provenance.}
Each run creates its own directory (an existing one is refused, never reused) with the git commit and dirty flag, the full configuration, all seeds, the pinned model revisions and adapter content hashes, the package list, GPU name, host and start time; step logs are append-only and a summary is written once. No results file is ever overwritten; the test split was opened once, on 2026-09-24, after an explicit approval marker was written
; 
no test episode or audit was rerun afterwards, and table-level $\kappa$  summaries were re-aggregated once from the stored audits to the per-game weighting stated above (Amendment A5).

\section{Pre-registered gates and amendments}
\label{app:gates}

The study was organised in stages with gates written before the corresponding data existed. Table~\ref{tab:gates} 
lists the gates and their verdicts; the amendments below were recorded in the specification, with their reasons, before the writer was trained (A1, A2), before the second reader was trained (A3) and before the test split was opened (A4); one was made after it (A5), which re-aggregates stored results and corrects one seed selection in the report script, and changes no run, audit or verdict.

\begin{table}[ht]
\caption{Pre-registered gates and verdicts. G1 was evaluated on validation games before training; G2 once, on the test split, with the trained writer fixed. Intervals are paired game bootstraps. G2b gives both the statistics as first computed and their correction (Amendment~A5).}
\label{tab:gates}
\begin{center}
\small
\setlength{\tabcolsep}{3pt}
\begin{tabular}{p{0.8cm} p{5.7cm} p{5.2cm} p{1.3cm}}
\toprule
gate & criterion & result & verdict \\
\midrule
G1a & $\kappa$ grows with lag: $\kappa(\Fp{=}40)-\kappa(\Fp{=}0)>0$, interval excluding 0, for the write-time formats & 1/3 formats on the primary GPU (plateau after ten rewrites), 3/3 on the replicate; in the measured-lag form $\kappa(\ell\ge5)-\kappa(\ell<5)=+1.08$ \ci{0.24}{1.97} & fail / pass ($\ell$ form) \\
G1b & the mean gap $\delta$ predicts online success across cells (Spearman) & $-0.84$ over 76 cells ($-0.90$ replicate) & pass \\
G1c & $\rho(S_{12},\text{return}) \ge 0.3$ and $\rho(S_{12})-\rho(S_1)>0$ on Phase-A write points & window form: $0.07$, fail; recursive form: $0.48$ \ci{0.39}{0.56} under the original protocol, $0.26$ \ci{0.13}{0.43} under $\Fp$ (91/200 write points without return variance) & fail (window) \\
\midrule
G2a & at $B{=}128$, $\Fp\in\{20,40\}$: \dssr $\ge$ best prompted $+8$\pts and $\ge$ \fmt{outcome} $+8$\pts (intervals $>0$); $\Fp{=}0$ not below $-3$\pts & $+1.5$ \ci{-0.4}{+3.3} over \fmt{summary}; $+1.5$ \ci{+0.2}{+3.0} over \fmt{outcome}; $\Fp{=}0$: $+7.0$ \ci{+1.9}{+12.2} & fail \\
G2b & $\kappa$ at $\Fp \in \{20, 40\}$ reduced by $\geq 40\,\%$ (interval $<0$);
      smaller $\kappa$-vs-$\Fp$ slope
    & as first computed: $1.456 \to 1.422$ ($-2\,\%$), difference $-0.03$
      $[-0.27, +0.20]$; slopes $0.022 \to 0.012$ per step (seed~0), difference
      $-0.010$ $[-0.022, +0.003]$. Corrected (A5): $1.50 \to 1.47$ ($-2\,\%$),
      $-0.03$ $[-0.27, +0.21]$; slopes $0.022 \to 0.022$ per step (three
      seeds), difference $+0.000$ $[-0.009, +0.009]$
    & fail \\
G2c & $K{=}12$ beats $K{=}1$ at $\Fp\in\{20,40\}$ (interval $>0$); the gap shrinks at $\Fp{=}0$ & $+1.5$ \ci{-0.4}{+3.3} at lag; $+3.7$ \ci{-3.3}{+10.4} at $\Fp{=}0$ & fail \\
G2d & under $R_2$, \dssr keeps $\ge 50$\,\% of its lag gain over \fmt{summary} & gain $+1.5$ \ci{-0.4}{+3.3} under $R_1$, $0.0$ under $R_2$ & fail (vacuous) \\
G2e & \fmt{no-hint} keeps $\ge 70$\,\% of \dssr's lag gain & $+1.5$ vs $+2.2$ \ci{0.0}{+5.0} & pass (vacuous) \\
plain & \dssr vs \fmt{guide} and \fmt{full} at lag, reported without threshold & $+0.9$ \ci{-0.9}{+2.8}; $-93.5$ \ci{-97.0}{-89.4} & reported \\
\bottomrule
\end{tabular}
\end{center}
\end{table}

\paragraph{Amendments.}
\emph{A1 (after the diagnostic).} The fixed-state (window) score was retired in favour of the recursive score, because the two were uncorrelated and only the recursive score predicted return; and the measured relevance lag $\ell$ replaced the pre-reveal filler $F$ as the lag variable of every claim, because $F$ did not move $\kappa$. \emph{A2 (after a causal pilot).} The filler walk was moved after the reveal (protocol $\Fp$); the whole diagnostic grid was re-run under $\Fp$ on both GPUs, and $\Fp$ is the protocol of every later stage. \emph{A3.} The rule ``one resident base model per GPU'', written for the 32\,GB card, was lifted on the 96\,GB card so that the Phi reader could be trained on reader-driven data written by the Qwen writer. \emph{A4.} A reachability-restricted sampling arm (\S\ref{sec:method}) was added after the credit-assignment finding, trained with the same collected episodes and starting adapter as the main line; gates are evaluated per line and the selected line is stated (it is A4). 
\emph{A5 (after the test evaluation; re-aggregation and one correction).}
No episode, audit or decision was rerun, and no gate verdict changes.
(i)~Table cells: the run summaries reported $\kappa$ per cell as the mean over
audited decision points; every table and figure now reports the mean of
per-game means stated in Appendix~\ref{app:metrics} 
(the two differ by up to 0.26~nats on the test split).
(ii)~Lag $\kappa$ at $\Fp \in \{20, 40\}$: the gate computation pooled each
game's audited points across the two lags before averaging over games; it is
now the mean over games of each game's two per-lag means, so that it equals the
average of the two cells of Table~\ref{tab:test}.
(iii)~The report script computed the \dssr{} slope in G2b from seed~0 alone,
while the $\kappa$ levels in the same gate pool the three seeds; the slope now
pools the three seeds as well. The seeds' slopes are 0.012, 0.031 and 0.023
nats per step; the \fmt{summary} slope, computed from per-(game,~$\Fp$) means
throughout, is unchanged.
For G2b, the statistics as first computed were $1.456 \to 1.422$ ($-2\,\%$),
difference $-0.03$ $[-0.27, +0.20]$, and slopes $0.022 \to 0.012$ per step
(seed~0), difference $-0.010$ $[-0.022, +0.003]$; corrected, they are
$1.50 \to 1.47$ ($-2\,\%$), $-0.03$ $[-0.27, +0.21]$, and slopes
$0.022 \to 0.022$ per step (three seeds), difference $+0.000$
$[-0.009, +0.009]$.

\section{Diagnostic: full tables}
\label{app:stage2}

\begin{table}[ht]
\caption{Validation grid under protocol $\Fp$ (primary GPU): success (\% of 45 games) at every budget, and mean $\kappa$ with game-bootstrap intervals at $B=128$ (200--270 audited decision points per cell). $\kappa$ is the mean of per-game means; $\beta$ is between $-0.14$ and $0.23$ in every cell.}
\label{tab:stage2full}
\begin{center}
\scriptsize
\setlength{\tabcolsep}{2.5pt}
\begin{tabular}{l l rrrr}
\toprule
format & quantity & $\Fp{=}0$ & $\Fp{=}10$ & $\Fp{=}20$ & $\Fp{=}40$ \\
\midrule
\fmt{full} & success & 97.8 & 97.8 & 97.8 & 97.8 \\
\fmt{oracle-b} & success, $B{=}64/128/256$ & 100 & 100 & 100 & 97.8 \\
\fmt{oracle-b} & $\kappa$, $B{=}128$ & $-0.19$ \ci{-0.30}{-0.08} & $-0.17$ \ci{-0.29}{-0.07} & $-0.21$ \ci{-0.29}{-0.13} & $-0.15$ \ci{-0.22}{-0.09} \\
\midrule
\fmt{lastk} & success, $B{=}64/128$ & 0 & 0 & 0 & 0 \\
 & success, $B{=}256$ & 2.2 & 0 & 0 & 0 \\
 & $\kappa$, $B{=}128$ & 0.71 \ci{0.54}{0.92} & 1.80 \ci{1.39}{2.22} & 1.46 \ci{1.10}{1.83} & 1.46 \ci{1.03}{1.92} \\
\fmt{summary} & success, $B{=}64$ & 0 & 0 & 0 & 0 \\
 & success, $B{=}128$ & 2.2 & 2.2 & 0 & 2.2 \\
 & success, $B{=}256$ & 20.0 & 6.7 & 2.2 & 2.2 \\
 & $\kappa$, $B{=}128$ & 0.85 \ci{0.55}{1.18} & 1.26 \ci{0.89}{1.70} & 1.47 \ci{1.07}{1.92} & 1.10 \ci{0.79}{1.46} \\
\fmt{slots} & success, $B{=}64$ & 0 & 0 & 0 & 0 \\
 & success, $B{=}128$ & 11.1 & 0 & 0 & 0 \\
 & success, $B{=}256$ & 15.6 & 4.4 & 2.2 & 0 \\
 & $\kappa$, $B{=}128$ & 0.49 \ci{0.33}{0.69} & 1.55 \ci{1.07}{2.10} & 1.38 \ci{0.98}{1.80} & 1.69 \ci{1.16}{2.31} \\
\fmt{belief} & success, $B{=}64$ & 0 & 0 & 0 & 0 \\
 & success, $B{=}128$ & 8.9 & 6.7 & 4.4 & 6.7 \\
 & success, $B{=}256$ & 28.9 & 15.6 & 8.9 & 6.7 \\
 & $\kappa$, $B{=}128$ & 0.79 \ci{0.49}{1.16} & 1.44 \ci{1.03}{1.91} & 1.27 \ci{0.92}{1.69} & 1.17 \ci{0.73}{1.67} \\
\fmt{guide} & success, $B{=}64$ & 0 & 0 & 0 & 0 \\
 & success, $B{=}128$ & 11.1 & 11.1 & 2.2 & 6.7 \\
 & success, $B{=}256$ & 40.0 & 31.1 & 31.1 & 22.2 \\
 & $\kappa$, $B{=}128$ & 0.66 \ci{0.40}{0.98} & 0.72 \ci{0.51}{0.94} & 0.84 \ci{0.58}{1.12} & 1.10 \ci{0.71}{1.56} \\
\bottomrule
\end{tabular}
\end{center}
\end{table}

\begin{table}[ht]
\caption{$\kappa$ (nats, mean of per-game means) at $B=64$ and $B=256$ under protocol $\Fp$ (validation, primary GPU).}
\label{tab:stage2kappaB}
\begin{center}
\small
\begin{tabular}{l rrrr rrrr}
\toprule
& \multicolumn{4}{c}{$B{=}64$} & \multicolumn{4}{c}{$B{=}256$} \\
\cmidrule(lr){2-5}\cmidrule(lr){6-9}
$\Fp$ & 0 & 10 & 20 & 40 & 0 & 10 & 20 & 40 \\
\midrule
\fmt{oracle-b} & $-0.37$ & $-0.22$ & $-0.20$ & $-0.31$ & $-0.26$ & $-0.15$ & $-0.18$ & $-0.25$ \\
\fmt{lastk} & 0.63 & 1.52 & 1.48 & 1.32 & 0.77 & 1.95 & 1.65 & 1.36 \\
\fmt{summary} & 0.93 & 1.42 & 1.29 & 1.56 & 0.42 & 1.31 & 1.50 & 1.50 \\
\fmt{slots} & 0.65 & 1.72 & 1.60 & 1.53 & 0.39 & 1.92 & 1.47 & 1.82 \\
\fmt{belief} & 0.70 & 1.24 & 1.11 & 1.45 & 0.50 & 1.15 & 1.05 & 0.95 \\
\fmt{guide} & 0.77 & 1.49 & 1.51 & 1.43 & 0.47 & 0.80 & 0.80 & 0.60 \\
\bottomrule
\end{tabular}
\end{center}
\end{table}

\begin{figure}[ht]
\begin{center}
\includegraphics[width=0.72\linewidth]{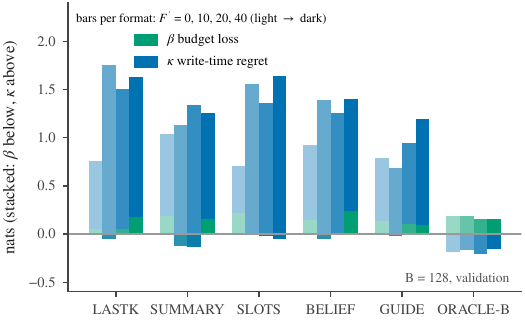}
\end{center}
\caption{Decomposition $\delta=\beta+\kappa$ per format at $B=128$ on validation games under protocol $\Fp$ (one bar per $\Fp$, light to dark): the lower segment is the budget loss $\beta$, the upper the write-time regret $\kappa$; \fmt{oracle-b} has negative $\kappa$.}
\label{fig:decomp}
\end{figure}

\begin{figure}[ht]
\begin{center}
\includegraphics[width=0.6\linewidth]{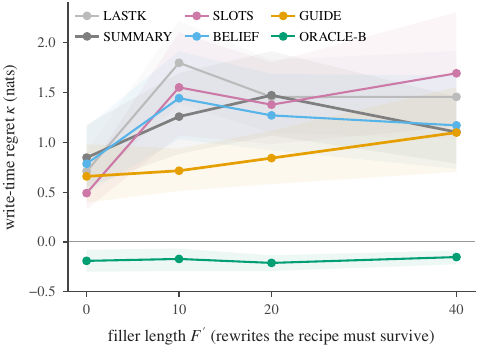}
\end{center}
\caption{Write-time regret $\kappa$ at $B=128$ against the post-reveal filler length $\Fp$ on validation games (game-bootstrap bands): $\kappa$ rises for every writer format within the first ten rewrites and stays high, while \fmt{oracle-b} stays below zero.}
\label{fig:f02}
\end{figure}

\paragraph{Decomposition and the original protocol.}
Figure~\ref{fig:decomp} shows $\delta=\beta+\kappa$ per cell; $\kappa$ dominates in every budgeted LLM cell. Under the original protocol (filler before the reveal) the $\kappa$-versus-filler curves were flat or non-monotone: the $F=40$ minus $F=0$ contrast was $-0.20$, $-0.34$ and $+0.24$ nats for \fmt{summary}, \fmt{belief} and \fmt{slots} at $B=128$ (no consistent sign on either GPU), and leave-one-chunk-out attributed the largest loss increase to the cookbook block at 87\,\% of decision points. Figure~\ref{fig:f02} shows the curves under protocol $\Fp$.

\begin{table}[ht]
\caption{$\kappa$, $\beta$ and the full-history loss against the measured relevance lag $\ell$ (validation, $B=128$ write-time formats, 900 sampled decision points, 532 with a defined $\ell$ at the 0.5-nat threshold; game-bootstrap intervals). The two other thresholds give the same picture. The chunk whose removal hurts most is the cookbook at 441 of the 532 points (83\,\%), a Phase-A room visit at 78 (15\,\%) and another Phase-B block at 13.}
\label{tab:ell}
\begin{center}
\small
\begin{tabular}{l r rrr rr}
\toprule
$\ell$ bin & $n$ & $\kappa$ & $\beta$ & $\NLL(h)$ & $\kappa$ (thr.\ 0.3) & $\kappa$ (thr.\ 1.0) \\
\midrule
1--4 & 107 & 0.61 \ci{0.38}{0.84} & 0.30 \ci{-0.00}{0.59} & 0.10 & 0.60 & 0.56 \\
5--9 & 15 & 1.89 \ci{0.33}{3.78} & $-1.61$ \ci{-3.76}{0.32} & 2.58 & 1.35 & 2.26 \\
10--19 & 123 & 1.97 \ci{1.41}{2.62} & 0.01 \ci{-0.27}{0.24} & 0.37 & 1.89 & 2.21 \\
20--39 & 146 & 1.87 \ci{1.41}{2.36} & 0.29 \ci{-0.01}{0.57} & 0.32 & 1.87 & 2.12 \\
40+ & 141 & 2.30 \ci{1.75}{2.92} & $-0.22$ \ci{-0.67}{0.15} & 0.68 & 2.18 & 2.50 \\
\bottomrule
\end{tabular}
\end{center}
\end{table}

\begin{figure}[ht]
\begin{center}
\includegraphics[width=0.6\linewidth]{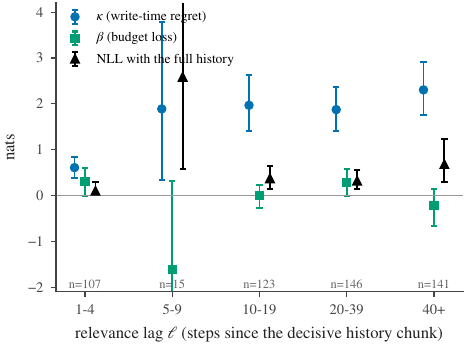}
\end{center}
\caption{$\kappa$, $\beta$ and the full-history loss per relevance-lag bin (Table~\ref{tab:ell}).}
\label{fig:ell}
\end{figure}

\begin{table}[ht]
\caption{Protocol pilot (\fmt{summary}, $B=128$, 45 validation games, replicate GPU): mean $\kappa$ under three placements of the same 40-step walk, and paired contrasts with intervals. ``Brief'' suppresses the re-printed room contents during the walk; ``post'' places the walk after the reveal.}
\label{tab:pilot}
\begin{center}
\small
\begin{tabular}{l r}
\toprule
condition & mean $\kappa$ \\
\midrule
baseline, $F{=}0$ & 0.69 \\
baseline, $F{=}40$ (walk before the reveal) & 0.45 \\
brief, $F{=}40$ & 0.92 \\
post, 40 steps after the reveal & 1.76 \\
post and brief & 1.13 \\
\midrule
contrast & estimate \ci{lo}{hi} \\
\midrule
brief $F{=}40$ $-$ baseline $F{=}40$ & $+0.47$ \ci{0.10}{0.93} \\
post $-$ baseline $F{=}0$ & $+1.07$ \ci{0.42}{1.75} \\
post $-$ baseline $F{=}40$ & $+1.31$ \ci{0.75}{1.92} \\
post and brief $-$ baseline $F{=}0$ & $+0.44$ \ci{-0.01}{0.92} \\
brief: $\kappa(F{=}40)-\kappa(F{=}0)$ & $+0.23$ \ci{-0.13}{0.65} \\
\bottomrule
\end{tabular}
\end{center}
\end{table}

\paragraph{Ties.}
$a^*$ is one optimal action and ties exist; on 500 sampled steps where the reader's argmax differs from $a^*$, the reader's action also shortens the oracle plan in 8.8\,\% of the cases, so ties add noise rather than bias to $\delta$ and $\kappa$, which are differences under the same $a^*$.

\section{Training the writer: data, hyper-parameters and round-by-round results}
\label{app:stage3}

\paragraph{Data per round.}
480 episodes on the 240 training games (two per game) with the current writer at temperature 0.7 and the frozen reader $R_1$ acting $\epsilon$-greedily ($\epsilon=0.1$), $\Fp \sim U\{0,10,20,40\}$ and $B \sim U\{64,128,256\}$; 3{,}984 write points sampled per round, half from Phase A, evenly over the ($\Fp$, $B$) cells, with a per-round seed; six candidates per write point at temperature 1.0 (the $i$-th of $n$ samples is seeded with seed$+i$, so candidate sets are reproducible); the recursive score for $K \in \{1,4,12\}$ from one set of roll-forwards. Under $\epsilon$-greedy the untrained writer wins 4.0\,\% of the training-game episodes. In the round-0 candidates 34\,\% of the states contain an admissible command verbatim. The A4 variant keeps 2{,}529 of the 3{,}984 write points (every write point at or before the reveal, and the post-reveal points whose input state holds a recipe directive) and writes 12 candidates at temperature 1.1 (5.4 GPU-hours of scoring per round). \fmt{outcome} scores 1{,}000 write points per round (at most 25\,\% from Phase A) by the normalised score of the episode continued from the candidate with the current writer at temperature 0 and the argmax reader; only 23--27\,\% of its write points have a return gap of at least 0.2 between candidates, so its rounds use 370--600 pairs.

\paragraph{Pairs and training.}
Pairs are best versus worst by $S_{12}$ with a gap of at least 0.3 nats, at most two per write point: 4{,}460 (round 1, 69\,\% of write points with a qualifying pair, median gap 0.58), 4{,}030 (round 2, 63\,\%) and 3{,}413 (round 3, 54\,\%, median gap 0.36) for the main line; 3{,}923 (83\,\%, median gap 0.96), 3{,}861 (median 0.88) and 3{,}980 (79\,\%, median 0.92) for the A4 variant, whose pair signal does not thin as the policy sharpens. Round 0 is one epoch of supervised fine-tuning of a fresh LoRA adapter (rank 32, all linear layers, learning rate $10^{-4}$) on the best candidate; rounds 1--3 are DPO with $\beta_{\mathrm{DPO}}=0.1$ plus $0.5\times$ the negative log-likelihood of the chosen state, learning rate $10^{-5}$, two epochs, effective batch 32, bf16, from the previous round's adapter as initialisation and reference (240--250 optimiser steps, about 40 minutes and 27\,GB peak per round on the 96\,GB GPU). No training example exceeds 895 tokens. The $K{=}1$, $K{=}4$, \fmt{no-hint} and \fmt{outcome} variants are paired from each round's candidates and trained with the identical recipe; \fmt{no-hint} rejects candidates that contain an admissible command verbatim before pairing. Evaluation after every round: online success and $\kappa$ on the 45 validation games with the \fmt{summary} prompt at $B=128$ and every $\Fp$.

\begin{table}[ht]
\caption{Validation success (\%) per writer, round and lag ($B=128$; round 0 = SFT, rounds 1--3 = DPO). Untrained \fmt{summary}: 2.2 / 2.2 / 0 / 2.2; \fmt{guide}: 11.1 / 11.1 / 2.2 / 6.7.}
\label{tab:rounds_success}
\begin{center}
\small
\begin{tabular}{l l rrrr r}
\toprule
writer & round & $\Fp{=}0$ & 10 & 20 & 40 & mean \\
\midrule
\dssr ($K{=}12$) & 0 & 13.3 & 0 & 2.2 & 0 & 3.9 \\
 & 1 & 11.1 & 4.4 & 2.2 & 2.2 & 5.0 \\
 & 2 & 4.4 & 8.9 & 4.4 & 0 & 4.4 \\
 & 3 & 13.3 & 0 & 0 & 2.2 & 3.9 \\
$K{=}4$ & 0 & 6.7 & 2.2 & 0 & 0 & 2.2 \\
 & 1 & 4.4 & 2.2 & 0 & 0 & 1.7 \\
 & 2 & 4.4 & 6.7 & 2.2 & 2.2 & 3.9 \\
 & 3 & 8.9 & 8.9 & 8.9 & 0 & 6.7 \\
$K{=}1$ & 0 & 4.4 & 4.4 & 2.2 & 0 & 2.8 \\
 & 1 & 17.8 & 6.7 & 4.4 & 0 & 7.2 \\
 & 2 & 8.9 & 0 & 2.2 & 0 & 2.8 \\
 & 3 & 20.0 & 2.2 & 0 & 2.2 & 6.1 \\
\fmt{no-hint} & 0 & 8.9 & 0 & 0 & 2.2 & 2.8 \\
 & 1 & 6.7 & 2.2 & 0 & 0 & 2.2 \\
 & 2 & 6.7 & 4.4 & 4.4 & 0 & 3.9 \\
 & 3 & 13.3 & 4.4 & 0 & 0 & 4.4 \\
\fmt{outcome} & 0 & 4.4 & 2.2 & 2.2 & 0 & 2.2 \\
 & 1 & 8.9 & 4.4 & 2.2 & 0 & 3.9 \\
 & 2 & 8.9 & 2.2 & 2.2 & 2.2 & 3.9 \\
 & 3 & 11.1 & 4.4 & 0 & 2.2 & 4.4 \\
A4 (selected line) & 1 & 4.4 & 0 & 0 & 6.7 & 2.8 \\
 & 2 & \textbf{26.7} & \textbf{11.1} & 4.4 & 2.2 & \textbf{11.1} \\
 & 3 & 15.6 & 2.2 & 2.2 & 2.2 & 5.6 \\
\bottomrule
\end{tabular}
\end{center}
\end{table}

\begin{table}[ht]
\caption{Validation $\kappa$ (nats, mean of per-game means) per writer, round and lag ($B=128$). Untrained \fmt{summary}: 0.85 / 1.26 / 1.47 / 1.10 (mean 1.17).}
\label{tab:rounds_kappa}
\begin{center}
\small
\begin{tabular}{l l rrrr r}
\toprule
writer & round & $\Fp{=}0$ & 10 & 20 & 40 & mean \\
\midrule
\dssr ($K{=}12$) & 0 & 0.71 & 1.24 & 1.17 & 1.70 & 1.20 \\
 & 1 & 0.52 & 1.60 & 1.59 & 1.87 & 1.40 \\
 & 2 & 0.47 & 1.06 & 1.17 & 1.41 & 1.02 \\
 & 3 & 0.79 & 1.11 & 1.02 & 1.14 & 1.01 \\
$K{=}4$ & 0 & 0.94 & 1.49 & 1.51 & 1.36 & 1.33 \\
 & 1 & 0.80 & 1.30 & 1.44 & 1.54 & 1.27 \\
 & 2 & 0.63 & 1.19 & 1.24 & 1.42 & 1.12 \\
 & 3 & 0.86 & 1.08 & 0.91 & 1.13 & 0.99 \\
$K{=}1$ & 0 & 0.69 & 1.40 & 1.33 & 1.38 & 1.20 \\
 & 1 & 0.76 & 1.56 & 1.25 & 1.52 & 1.28 \\
 & 2 & 0.81 & 1.34 & 1.26 & 1.15 & 1.14 \\
 & 3 & 0.44 & 1.15 & 1.24 & 1.57 & 1.10 \\
\fmt{no-hint} & 0 & 0.68 & 1.16 & 1.28 & 1.08 & 1.05 \\
 & 1 & 0.67 & 1.06 & 1.24 & 1.11 & 1.02 \\
 & 2 & 0.75 & 1.13 & 1.50 & 1.32 & 1.18 \\
 & 3 & 0.53 & 1.30 & 1.41 & 1.43 & 1.17 \\
\fmt{outcome} & 0 & 0.75 & 1.92 & 1.46 & 1.67 & 1.45 \\
 & 1 & 0.80 & 1.23 & 1.28 & 1.59 & 1.22 \\
 & 2 & 0.93 & 1.61 & 1.37 & 1.44 & 1.34 \\
 & 3 & 0.62 & 1.30 & 1.26 & 1.38 & 1.14 \\
A4 (selected line) & 1 & 0.61 & 1.40 & 1.36 & 1.51 & 1.22 \\
 & 2 & 0.61 & 1.13 & 1.19 & 1.46 & 1.10 \\
 & 3 & 0.58 & 1.05 & 1.15 & 1.33 & 1.03 \\
\bottomrule
\end{tabular}
\end{center}
\end{table}

\paragraph{Game-paired $\kappa$ contrasts against the untrained writer (validation).}
Relative change of the game-mean $\kappa$ against the untrained writer, the ratio recomputed in every game resample; ``lag'' is the mean of a game's $\Fp{=}20$ and $\Fp{=}40$ means. Main line round 1: $\Fp{=}0$ $-38$\,\% \ci{-68}{+2}, lag $+34$\,\% \ci{+4}{+75}. Round 2: $\Fp{=}0$ $-45$\,\% \ci{-66}{-11}, lag $0$\,\% \ci{-21}{+25}. Round 3: $\Fp{=}20$ $-31$\,\% \ci{-53}{-1}, lag $-16$\,\% \ci{-34}{+7}; $K{=}4$ round 3: $\Fp{=}20$ $-39$\,\% \ci{-57}{-14}, lag $-21$\,\% \ci{-40}{+2}. A4 round 2, seed 0: $\Fp{=}0$ $-28$\,\% \ci{-61}{+21}, lag $+3$\,\% \ci{-18}{+29}; seed 1: $-39$\,\% \ci{-61}{-3} and $-16$\,\% \ci{-40}{+15}; seed 2: $-64$\,\% \ci{-83}{-32} and $+4$\,\% \ci{-18}{+33}; three seeds pooled: lag $-3$\,\% \ci{-19}{+20}.

\begin{table}[ht]
\caption{Credit assignment on the training candidates (post-reveal write points). Coverage = fraction of the recipe's verb--ingredient directives present in a state; ``all zero'' = no candidate holds a directive; ``differ'' = candidates differ in coverage; the last two columns are, among differing points, how often the best candidate by $S_{12}$ has higher (lower) coverage than the worst.}
\label{tab:credit}
\begin{center}
\scriptsize
\setlength{\tabcolsep}{2.5pt}
\begin{tabular}{l l r rrr rr rr}
\toprule
candidates & group & $n$ & cov.\ all & cov.\ best & cov.\ worst & all zero & differ & best$>$worst & best$<$worst \\
\midrule
main, round 0 & all & 2818 & 0.24 & 0.23 & 0.24 & 0.51 & 0.09 & 0.28 & 0.29 \\
 & $\Fp{=}0$ & 439 & 0.44 & 0.44 & 0.43 & 0.19 & 0.20 & 0.24 & 0.33 \\
 & $\Fp{=}10$ & 632 & 0.30 & 0.30 & 0.31 & 0.41 & 0.10 & 0.29 & 0.33 \\
 & $\Fp{=}20$ & 760 & 0.20 & 0.19 & 0.20 & 0.58 & 0.07 & 0.30 & 0.26 \\
 & $\Fp{=}40$ & 987 & 0.13 & 0.13 & 0.13 & 0.67 & 0.05 & 0.31 & 0.18 \\
 & 1--4 steps after reveal & 331 & 0.42 & 0.42 & 0.41 & 0.27 & 0.25 & 0.28 & 0.27 \\
 & 5--14 & 639 & 0.29 & 0.29 & 0.30 & 0.46 & 0.10 & 0.22 & 0.33 \\
 & 15--29 & 749 & 0.21 & 0.21 & 0.20 & 0.54 & 0.07 & 0.35 & 0.26 \\
 & 30+ & 1099 & 0.17 & 0.16 & 0.17 & 0.59 & 0.05 & 0.27 & 0.29 \\
main, round 1 & all & 2799 & 0.25 & 0.25 & 0.24 & 0.51 & 0.10 & 0.35 & 0.26 \\
 & 1--4 steps after reveal & 348 & 0.49 & 0.50 & 0.47 & 0.23 & 0.27 & 0.40 & 0.29 \\
A4, round 1 & all & 1344 & 0.50 & 0.51 & 0.49 & 0.01 & 0.29 & 0.31 & 0.22 \\
A4, round 2 & all & 1412 & 0.53 & 0.54 & 0.52 & 0.01 & 0.27 & 0.27 & 0.16 \\
 & 1--4 steps after reveal & 269 & 0.67 & 0.71 & 0.65 & 0.01 & 0.43 & 0.38 & 0.14 \\
\bottomrule
\end{tabular}
\end{center}
\end{table}
\begin{figure}[ht]
\begin{center}
\includegraphics[width=0.7\linewidth]{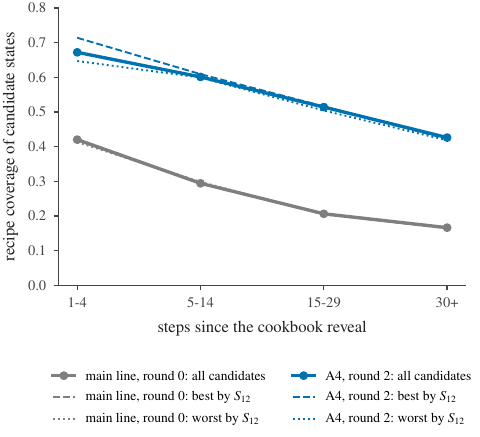}
\end{center}
\caption{Recipe coverage of all candidates, of the best and of the worst by $S_{12}$ against steps since the reveal (main line round 0, A4 round 2): the score does not select for retention, and coverage decays with rewrites.}
\label{fig:f07}
\end{figure}

\paragraph{Selection, seeds and budget.}
Selection (online success at $B=128$ averaged over $\Fp$, validation): A4 round 2, 11.1\,\%; $K{=}1$ round 1, 7.2\,\%; $K{=}4$ round 3, 6.7\,\%; main-line \dssr round 1, 5.0\,\%; \fmt{no-hint} round 3 and \fmt{outcome} round 3, 4.4\,\%. The best round of each variant is its test representative. The selected configuration was retrained from the same initialisation and pairs with two more seeds (Table~\ref{tab:seeds}). Budget sweep of seed 0 on validation: $B{=}64$ 4.4 / 0 / 0 / 0\,\% (untrained 0 / 0 / 0 / 0); $B{=}256$ 37.8 / 24.4 / 15.6 / 15.6\,\% (untrained 20.0 / 6.7 / 2.2 / 2.2), so on validation the trained writer's gain reaches every lag once the state has room.

\begin{table}[ht]
\caption{Seeds of the selected configuration (A4 round 2) on validation, $B=128$: success (\%) and $\kappa$ (mean of per-game means) at $\Fp=0/10/20/40$.}
\label{tab:seeds}
\begin{center}
\small
\begin{tabular}{l rrrr rrrr}
\toprule
& \multicolumn{4}{c}{success} & \multicolumn{4}{c}{$\kappa$} \\
\cmidrule(lr){2-5}\cmidrule(lr){6-9}
seed & 0 & 10 & 20 & 40 & 0 & 10 & 20 & 40 \\
\midrule
0 (selected) & 26.7 & 11.1 & 4.4 & 2.2 & 0.61 & 1.13 & 1.19 & 1.46 \\
1 & 8.9 & 11.1 & 2.2 & 2.2 & 0.51 & 1.26 & 0.91 & 1.26 \\
2 & 22.2 & 8.9 & 8.9 & 4.4 & 0.31 & 1.34 & 1.14 & 1.55 \\
mean $\pm$ sd & 19.3 $\pm$ 7.6 & 10.4 $\pm$ 1.0 & 5.2 $\pm$ 2.8 & 3.0 $\pm$ 1.0 & 0.48 & 1.24 & 1.08 & 1.42 \\
untrained & 2.2 & 2.2 & 0 & 2.2 & 0.85 & 1.26 & 1.47 & 1.10 \\
\bottomrule
\end{tabular}
\end{center}
\end{table}

\clearpage

\section{Test split: intervals, transfer reader, budget frontier, rewrite bins}
\label{app:stage4}

\begin{table}[H]
\caption{Test split, $B=128$: success (\%) with game-bootstrap intervals per lag, under $R_1$ and $R_2$; the last column is the lag $\kappa$ at $\Fp\in\{20,40\}$ (mean over games of a game's mean of its $\Fp{=}20$ and $\Fp{=}40$ means, so it equals the average of the two cells of Table~\ref{tab:test}) with its interval.}
\label{tab:testci}
\begin{center}
\scriptsize
\setlength{\tabcolsep}{2pt}
\begin{tabular}{l l rrrr r}
\toprule
reader & condition & $\Fp{=}0$ & 10 & 20 & 40 & $\kappa$, $\Fp\in\{20,40\}$ \\
\midrule
$R_1$ & \fmt{full} & 95.6 \ci{91.1}{98.9} & 96.7 \ci{92.2}{100} & 96.7 \ci{92.2}{100} & 94.4 \ci{88.9}{98.9} & -- \\
 & \fmt{oracle-b} & 98.9 \ci{96.7}{100} & 100 & 98.9 \ci{96.7}{100} & 97.8 \ci{94.4}{100} & -- \\
 & \fmt{lastk} & 0 & 0 & 0 & 0 & -- \\
 & \fmt{summary} & 3.3 \ci{0.0}{7.8} & 2.2 \ci{0.0}{5.6} & 1.1 \ci{0.0}{3.3} & 0 & 1.50 \ci{1.23}{1.80} \\
 & \fmt{belief} & 11.1 \ci{5.6}{17.8} & 0 & 0 & 0 & 1.64 \ci{1.36}{1.93} \\
 & \fmt{slots} & 10.0 \ci{4.4}{16.7} & 0 & 0 & 0 & 1.80 \ci{1.50}{2.12} \\
 & \fmt{guide} & 14.4 \ci{7.8}{22.2} & 6.7 \ci{2.2}{12.2} & 0 & 2.2 \ci{0.0}{5.6} & 1.11 \ci{0.89}{1.34} \\
 & \dssr seed 0 & 3.3 \ci{0.0}{7.8} & 1.1 \ci{0.0}{3.3} & 1.1 \ci{0.0}{3.3} & 0 & 1.38 \ci{1.16}{1.62} \\
 & \dssr seed 1 & 15.6 \ci{8.9}{23.3} & 6.7 \ci{2.2}{12.2} & 5.6 \ci{1.1}{11.1} & 2.2 \ci{0.0}{5.6} & 1.50 \ci{1.25}{1.76} \\
 & \dssr seed 2 & 12.2 \ci{5.6}{20.0} & 3.3 \ci{0.0}{7.8} & 0 & 3.3 \ci{0.0}{7.8} & 1.52 \ci{1.24}{1.82} \\
 & \dssr main line (r1) & 7.8 \ci{3.3}{13.3} & 1.1 \ci{0.0}{3.3} & 1.1 \ci{0.0}{3.3} & 1.1 \ci{0.0}{3.3} & 1.68 \ci{1.39}{1.99} \\
 & \fmt{outcome} (r3) & 8.9 \ci{3.3}{15.6} & 0 & 1.1 \ci{0.0}{3.3} & 0 & 1.68 \ci{1.39}{1.98} \\
 & $K{=}1$ (r1) & 6.7 \ci{2.2}{12.2} & 0 & 1.1 \ci{0.0}{3.3} & 0 & 1.60 \ci{1.31}{1.90} \\
 & $K{=}4$ (r3) & 11.1 \ci{5.6}{17.8} & 1.1 \ci{0.0}{3.3} & 2.2 \ci{0.0}{5.6} & 1.1 \ci{0.0}{3.3} & 1.22 \ci{0.98}{1.48} \\
 & \fmt{no-hint} (r3) & 8.9 \ci{3.3}{15.6} & 0 & 3.3 \ci{0.0}{7.8} & 2.2 \ci{0.0}{5.6} & 1.30 \ci{1.07}{1.56} \\
\midrule
$R_2$ & \fmt{full} & 77.8 \ci{68.9}{85.6} & 64.4 \ci{54.4}{74.4} & 64.4 \ci{54.4}{74.4} & 67.8 \ci{57.8}{77.8} & -- \\
 & \fmt{guide} & 11.1 \ci{5.6}{17.8} & 5.6 \ci{1.1}{11.1} & 3.3 \ci{0.0}{7.8} & 2.2 \ci{0.0}{5.6} & -- \\
 & \fmt{summary} & 2.2 \ci{0.0}{5.6} & 1.1 \ci{0.0}{3.3} & 0 & 0 & -- \\
 & \dssr seed 0 & 5.6 \ci{1.1}{11.1} & 1.1 \ci{0.0}{3.3} & 0 & 0 & -- \\
\bottomrule
\end{tabular}
\end{center}
\end{table}

\begin{table}[H]
\caption{Budget frontier on the test split (replicate GPU): success (\%) at $\Fp = 0/10/20/40$ for $B \in \{64,128,256\}$. \fmt{full}: 95.6 / 95.6 / 96.7 / 94.4; \fmt{oracle-b}: 98.9 / 98.9 / 98.9 / 97.8 at every budget.}
\label{tab:frontier}
\begin{center}
\small
\setlength{\tabcolsep}{4pt}
\begin{tabular}{l rrrr rrrr rrrr}
\toprule
& \multicolumn{4}{c}{$B{=}64$} & \multicolumn{4}{c}{$B{=}128$} & \multicolumn{4}{c}{$B{=}256$} \\
\cmidrule(lr){2-5}\cmidrule(lr){6-9}\cmidrule(lr){10-13}
$\Fp$ & 0 & 10 & 20 & 40 & 0 & 10 & 20 & 40 & 0 & 10 & 20 & 40 \\
\midrule
\fmt{summary} & 0 & 0 & 0 & 0 & 7.8 & 0 & 0 & 1.1 & 7.8 & 1.1 & 2.2 & 1.1 \\
\fmt{belief} & 2.2 & 0 & 1.1 & 1.1 & 16.7 & 1.1 & 1.1 & 1.1 & 16.7 & 6.7 & 8.9 & 2.2 \\
\fmt{slots} & 0 & 0 & 0 & 0 & 12.2 & 0 & 0 & 0 & 20.0 & 2.2 & 0 & 0 \\
\fmt{guide} & 0 & 0 & 0 & 0 & 16.7 & 2.2 & 5.6 & 2.2 & \textbf{30.0} & \textbf{16.7} & \textbf{15.6} & \textbf{15.6} \\
\dssr (A4 r2, seed 0) & 2.2 & 0 & 0 & 0 & 14.4 & 4.4 & 4.4 & 1.1 & 18.9 & 14.4 & 2.2 & 5.6 \\
\dssr main line (r1) & 1.1 & 0 & 0 & 0 & 7.8 & 3.3 & 0 & 0 & 17.8 & 5.6 & 8.9 & 4.4 \\
\fmt{outcome} (r3) & 2.2 & 0 & 0 & 0 & 7.8 & 3.3 & 0 & 0 & 15.6 & 5.6 & 2.2 & 1.1 \\
\bottomrule
\end{tabular}
\end{center}
\end{table}

\begin{figure}[H]
\begin{center}
\includegraphics[width=0.55\linewidth]{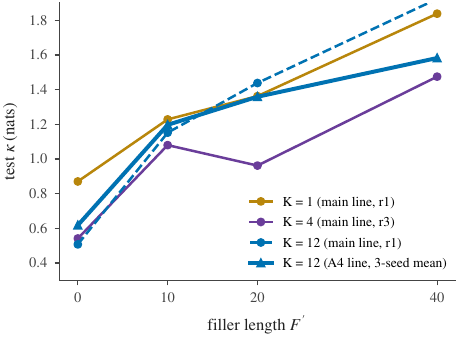}
\end{center}
\caption{Horizon ablation on the test split: $\kappa$ against $\Fp$ for writers trained with $K=1$, $4$ and $12$ (the $K=12$ A4 line is the three-seed mean).}
\label{fig:horizon}
\end{figure}

\begin{figure}[H]
\begin{center}
\includegraphics[width=0.62\linewidth]{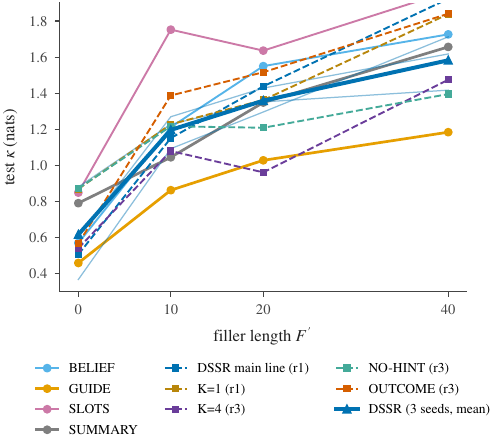}
\end{center}
\caption{$\kappa$ against $\Fp$ on the test split ($B=128$) for the untrained formats, the three \dssr seeds (thin) with their mean (thick) and the variant writers.}
\label{fig:kappa_before_after}
\end{figure}
\begin{figure}[ht]
\begin{center}
\includegraphics[width=0.6\linewidth]{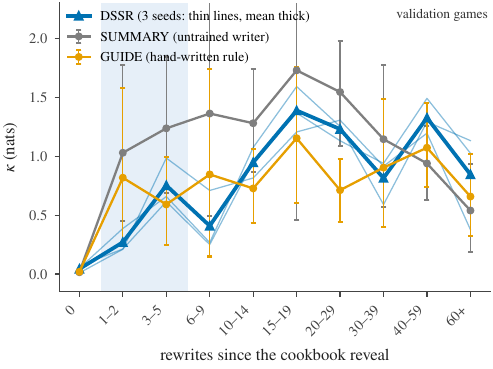}
\end{center}
\caption{Mean $\kappa$ by rewrites since the reveal on validation games (Table~\ref{tab:rewrites}); shaded: 1--5 rewrites.}
\label{fig:f08}
\end{figure}
\begin{figure}[ht]
\begin{center}
\includegraphics[width=0.6\linewidth]{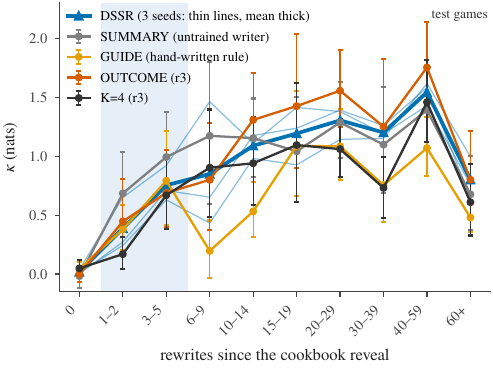}
\end{center}
\caption{Mean $\kappa$ by rewrites since the reveal on the test games (Table~\ref{tab:rewrites}).}
\label{fig:f09}
\end{figure}

\begin{table}[H]
\caption{Mean $\kappa$ (nats) by the number of rewrites since the cookbook reveal (audited decision points, validation and test, $B=128$); the number of points per cell is in Table~\ref{tab:rewrites_n}.}
\label{tab:rewrites}
\begin{center}
\scriptsize
\setlength{\tabcolsep}{4pt}
\begin{tabular}{l rrrrrrrrrr}
\toprule
condition & 0 & 1--2 & 3--5 & 6--9 & 10--14 & 15--19 & 20--29 & 30--39 & 40--59 & 60+ \\
\midrule
\fmt{summary}, valid & 0.02 & 1.03 & 1.24 & 1.36 & 1.28 & 1.73 & 1.54 & 1.14 & 0.94 & 0.54 \\
\dssr seed 0, valid & 0.05 & 0.39 & 0.66 & 0.27 & 1.07 & 1.59 & 1.25 & 0.59 & 1.29 & 1.13 \\
\dssr seed 1, valid & 0.01 & 0.21 & 0.98 & 0.71 & 0.82 & 1.21 & 1.31 & 0.91 & 1.20 & 0.38 \\
\dssr seed 2, valid & 0.07 & 0.21 & 0.61 & 0.25 & 0.95 & 1.37 & 1.13 & 0.94 & 1.49 & 1.03 \\
\fmt{guide}, valid & 0.02 & 0.82 & 0.59 & 0.85 & 0.73 & 1.15 & 0.71 & 0.90 & 1.07 & 0.66 \\
\midrule
\fmt{summary}, test & $-0.02$ & 0.68 & 0.99 & 1.17 & 1.15 & 1.04 & 1.29 & 1.10 & 1.39 & 0.68 \\
\dssr seed 0, test & 0.04 & 0.65 & 0.92 & 1.46 & 1.11 & 1.41 & 1.38 & 1.19 & 1.43 & 0.76 \\
\dssr seed 1, test & 0.02 & 0.24 & 0.63 & 0.43 & 0.98 & 0.93 & 1.14 & 1.15 & 1.62 & 1.00 \\
\dssr seed 2, test & 0.00 & 0.27 & 0.71 & 0.65 & 1.18 & 1.24 & 1.39 & 1.26 & 1.58 & 0.64 \\
\fmt{guide}, test & 0.05 & 0.38 & 0.79 & 0.20 & 0.53 & 1.09 & 1.08 & 0.75 & 1.07 & 0.48 \\
\fmt{outcome}, test & $-0.00$ & 0.45 & 0.69 & 0.80 & 1.31 & 1.43 & 1.56 & 1.25 & 1.75 & 0.80 \\
$K{=}4$, test & 0.05 & 0.17 & 0.67 & 0.90 & 0.94 & 1.09 & 1.06 & 0.73 & 1.46 & 0.61 \\
\bottomrule
\end{tabular}
\end{center}
\end{table}

\begin{table}[H]
\caption{Number of audited decision points per rewrite bin for Table~\ref{tab:rewrites}.}
\label{tab:rewrites_n}
\begin{center}
\scriptsize
\setlength{\tabcolsep}{4pt}
\begin{tabular}{l rrrrrrrrrr}
\toprule
condition & 0 & 1--2 & 3--5 & 6--9 & 10--14 & 15--19 & 20--29 & 30--39 & 40--59 & 60+ \\
\midrule
\fmt{summary}, valid & 39 & 67 & 45 & 20 & 152 & 29 & 183 & 44 & 229 & 67 \\
\dssr seed 0, valid & 26 & 58 & 50 & 43 & 152 & 54 & 182 & 47 & 245 & 65 \\
\dssr seed 1, valid & 29 & 66 & 53 & 37 & 139 & 56 & 196 & 54 & 246 & 75 \\
\dssr seed 2, valid & 27 & 57 & 51 & 45 & 153 & 56 & 182 & 64 & 230 & 83 \\
\fmt{guide}, valid & 33 & 59 & 44 & 32 & 125 & 46 & 187 & 59 & 260 & 50 \\
\midrule
\fmt{summary}, test & 69 & 132 & 130 & 69 & 247 & 85 & 364 & 107 & 493 & 151 \\
\dssr seed 0, test & 65 & 126 & 109 & 63 & 240 & 88 & 394 & 110 & 523 & 152 \\
\dssr seed 1, test & 62 & 124 & 115 & 83 & 263 & 93 & 398 & 119 & 519 & 138 \\
\dssr seed 2, test & 56 & 101 & 107 & 81 & 252 & 132 & 412 & 124 & 509 & 142 \\
\fmt{guide}, test & 64 & 117 & 92 & 63 & 231 & 114 & 367 & 136 & 547 & 176 \\
\fmt{outcome}, test & 61 & 119 & 105 & 71 & 240 & 102 & 412 & 120 & 528 & 123 \\
$K{=}4$, test & 51 & 108 & 122 & 94 & 257 & 101 & 379 & 125 & 550 & 140 \\
\bottomrule
\end{tabular}
\end{center}
\end{table}

\begin{table}[H]
\caption{Game-paired reduction of $\kappa$ against \fmt{summary} per rewrite bin: difference of per-game means (nats, positive = lower $\kappa$ than \fmt{summary}) with bootstrap interval, and the relative reduction. The three \dssr seeds are pooled. The 0-rewrite bin is omitted (baseline $\kappa \approx 0$).}
\label{tab:rewrites_contrast}
\begin{center}
\scriptsize
\begin{tabular}{l l rrrr}
\toprule
split & writer & bin & games & difference \ci{lo}{hi} & relative \\
\midrule
valid & \dssr (3 seeds) & 1--2 & 39 & $+0.83$ \ci{+0.20}{+1.65} & $-81$\,\% \\
 & & 3--5 & 30 & $+0.64$ \ci{+0.10}{+1.25} & $-52$\,\% \\
 & & 6--9 & 11 & $+0.60$ \ci{-0.14}{+1.63} & $-44$\,\% \\
 & & 10--14 & 44 & $+0.27$ \ci{-0.21}{+0.73} & $-21$\,\% \\
 & & 15--19 & 15 & $+0.22$ \ci{-1.13}{+1.65} & $-13$\,\% \\
 & & 20--29 & 43 & $+0.28$ \ci{-0.14}{+0.69} & $-18$\,\% \\
 & & 30--39 & 18 & $+0.02$ \ci{-0.68}{+0.68} & $-2$\,\% \\
 & & 40--59 & 45 & $-0.35$ \ci{-0.64}{-0.07} & $+38$\,\% \\
 & & 60+ & 19 & $-0.26$ \ci{-0.77}{+0.30} & $+50$\,\% \\
\midrule
test & \dssr (3 seeds) & 1--2 & 76 & $+0.21$ \ci{-0.13}{+0.60} & $-31$\,\% \\
 & & 3--5 & 72 & $+0.18$ \ci{-0.16}{+0.57} & $-19$\,\% \\
 & & 6--9 & 33 & $-0.04$ \ci{-0.57}{+0.46} & $+5$\,\% \\
 & & 10--14 & 78 & $+0.05$ \ci{-0.33}{+0.39} & $-4$\,\% \\
 & & 20--29 & 88 & $+0.03$ \ci{-0.24}{+0.33} & $-3$\,\% \\
 & & 40--59 & 89 & $-0.08$ \ci{-0.35}{+0.20} & $+6$\,\% \\
test & \fmt{guide} & 1--2 & 64 & $+0.39$ \ci{-0.01}{+0.81} & $-53$\,\% \\
 & & 3--5 & 55 & $+0.33$ \ci{-0.07}{+0.69} & $-32$\,\% \\
 & & 10--14 & 64 & $+0.54$ \ci{+0.10}{+0.98} & $-49$\,\% \\
 & & 20--29 & 84 & $+0.23$ \ci{-0.18}{+0.64} & $-17$\,\% \\
 & & 30--39 & 38 & $+0.45$ \ci{-0.15}{+1.10} & $-43$\,\% \\
 & & 40--59 & 89 & $+0.31$ \ci{+0.01}{+0.64} & $-23$\,\% \\
test & \fmt{outcome} & 1--2 & 63 & $+0.13$ \ci{-0.37}{+0.60} & $-22$\,\% \\
 & & 3--5 & 57 & $+0.18$ \ci{-0.31}{+0.67} & $-20$\,\% \\
 & & 20--29 & 85 & $-0.26$ \ci{-0.70}{+0.18} & $+20$\,\% \\
 & & 40--59 & 89 & $-0.37$ \ci{-0.68}{-0.06} & $+27$\,\% \\
\bottomrule
\end{tabular}
\end{center}
\end{table}

\section{Exploratory lag grid on validation}
\label{app:laggrid}

After the test evaluation we asked whether the trained writer's gain is confined to $\Fp=0$ or extends to a few rewrites, since the pre-registered grid has no point between 0 and 10. Additional post-reveal filler lengths $\Fp \in \{2,3,5,7,9,15,30\}$ were run on the \emph{validation} games only (the test split was not reopened) for the untrained formats, the three seeds of the selected writer, the main-line writer, \fmt{outcome} and $K{=}4$, with the same seeded walk, so the new lengths are prefixes of the pre-registered walk; a $B=256$ grid at $\Fp \in \{0,5,10,20,40\}$ was added for \fmt{summary}, \fmt{guide} and the three seeds. Every new cell was verified against the pre-registered runs: episodes are byte-identical over the shared prefix (exploration, reveal and the first $\Fp$ filler steps), the $B=256$ cells that repeat pre-registered lengths reproduce them episode by episode, every summary equals a recomputation from the raw logs, and every $\kappa$ audit passes its rescoring check. This is a post-hoc exploration: it enters no gate, and the lag-band wording of \S\ref{sec:main} rests on the rewrite-binned $\kappa$ analysis of the pre-registered runs, which the grid corroborates. Table~\ref{tab:laggrid_gain} gives the game-paired success gain of the selected writer over \fmt{summary}: present at every length up to 10, halved by 15--20 and gone at 30--40; \fmt{guide} (11.1 / 13.3 / 8.9 / 4.4 / 6.7 / 8.9 / 11.1 / 4.4 / 2.2 / 6.7 / 6.7\,\% at $\Fp = 0/2/3/5/7/9/10/15/20/30/40$) and $K{=}4$ (8.9 / 8.9 / 11.1 / 11.1 / 15.6 / 4.4 / 8.9 / 11.1 / 8.9 / 2.2 / 0) stay in the same band as the seeds. Two caveats for reading the raw curve. The maps are trees, so after an odd number of filler steps Phase B never starts in the kitchen, whereas after an even number it does in about half of the games (24 / 27 / 18 / 18 / 17 of 45 at $\Fp = 2/10/20/30/40$, all 45 at $\Fp=0$); episodes that start elsewhere are about twice as long ($\approx 25$ against $\approx 12$ Phase-B steps) and harder, and conditioning on the start room turns the bumps of the raw curve into two monotone decays (pooled seeds, kitchen starts: 19.3 $\to$ 12.5 $\to$ 13.6 $\to$ 7.4 $\to$ 3.7 $\to$ 2.0\,\% at $\Fp = 0, 2, 10, 20, 30, 40$; other starts: 11.1 $\to$ 9.6 $\to$ 9.6 $\to$ 8.9 $\to$ 6.7 $\to$ 5.6 $\to$ 6.7 $\to$ 3.7 $\to$ 1.2 $\to$ 3.6\,\% at $\Fp = 2, 3, 5, 7, 9, 10, 15, 20, 30, 40$); the remaining bumps are single-seed noise (one game is 2.2 points). And the mean $\kappa$ of a cell does not fall at the short lengths (three-seed means 0.73--1.06 nats at $\Fp = 2$--$9$ against 0.82--0.97 for \fmt{summary}) because it averages over all audited Phase-B points, most of which lie well beyond ten rewrites after the reveal whatever $\Fp$ is; the short-lag statistic is the rewrite-binned $\kappa$ of Table~\ref{tab:rewrites}. At $B=256$ the seeds' gain over \fmt{summary} holds at every length (31.9 / 17.0 / 19.3 / 14.1 / 12.6 against 20.0 / 8.9 / 6.7 / 2.2 / 2.2\,\% at $\Fp = 0/5/10/20/40$, consistently across seeds) while \fmt{guide} stays above (40.0 / 31.1 / 31.1 / 31.1 / 22.2).

\begin{table}[ht]
\caption{Success gain of the selected writer (three seeds pooled per game) over the untrained \fmt{summary} writer at every filler length of the exploratory grid (validation, $B=128$): percentage points with 95\,\% game-paired bootstrap intervals over the 45 games (10{,}000 resamples). Short and medium lag ($\Fp \le 9$): every interval excludes zero; $\Fp = 10$: interval touching zero; long lag (15--20): smaller but above zero; extra-long (30--40): not distinguishable from zero.}
\label{tab:laggrid_gain}
\begin{center}
\scriptsize
\setlength{\tabcolsep}{3.5pt}
\begin{tabular}{l rrrrrrrrrrr}
\toprule
$\Fp$ & 0 & 2 & 3 & 5 & 7 & 9 & 10 & 15 & 20 & 30 & 40 \\
\midrule
\dssr, 3-seed mean (\%) & 19.3 & 11.9 & 9.6 & 9.6 & 8.9 & 6.7 & 10.4 & 6.7 & 5.2 & 2.2 & 3.0 \\
\fmt{summary} (\%) & 2.2 & 2.2 & 0 & 0 & 0 & 0 & 2.2 & 0 & 0 & 0 & 2.2 \\
gain (points) & $+17.0$ & $+9.6$ & $+9.6$ & $+9.6$ & $+8.9$ & $+6.7$ & $+8.1$ & $+6.7$ & $+5.2$ & $+2.2$ & $+0.7$ \\
interval, lower & $+8.1$ & $+3.7$ & $+5.2$ & $+4.4$ & $+4.4$ & $+2.2$ & $-0.0$ & $+2.2$ & $+1.5$ & $+0.0$ & $-5.2$ \\
interval, upper & $+26.7$ & $+14.8$ & $+14.8$ & $+14.8$ & $+14.8$ & $+11.9$ & $+16.3$ & $+12.6$ & $+9.6$ & $+5.2$ & $+5.2$ \\
\bottomrule
\end{tabular}
\end{center}
\end{table}

\section{Qualitative examples}
\label{app:qual}

\paragraph{A recipe directive degrading across rewrites (\fmt{summary}, $B=128$, $\Fp=40$, validation game with 12 rooms; lost at step 81 with score 1/11).}
The cookbook, read at step 33, lists \emph{red bell pepper, carrot, banana} with the directions \emph{chop the red bell pepper; roast the red bell pepper; chop the carrot; roast the carrot; chop the banana; roast the banana; prepare meal}. The state written at the reveal step still holds the recipe, already re-ordered:

{\small
\begin{quote}
In kitchen: cookbook, ingredients (peppers, carrot, pork chop, apple, potato), knife, empty oven, stove. Secret button activated wall (reveals stove). Exits: east (frosted door), south (sliding patio door). Goal: cook meal using recipe. Recipe found: roast red bell pepper, carrot, banana; chop all; prepare meal. Pork chop, apple, potato not in recipe. Missing banana.
\end{quote}
}

After the 40 filler steps, at the first decision (step 73), the recipe is gone:

{\small
\begin{quote}
In driveway: east exit available (no door); need to reach kitchen for cookbook; last went north to living room; now at living room with open door south; should go north to find exit without door.
\end{quote}
}

At the fatal step (80) the reader, reading \emph{Carrying: carrot, knife, diced pork chop. Ingredients left: peppers, apple, potato. Goal: enjoy meal. Exits: east (frosted), south (sliding patio), west (doorless). Secret button activated. Score: +1.}, chose \texttt{cook carrot with stove} where $a^*$ was \texttt{take red bell pepper from fridge}; frying the carrot made the game unwinnable.

\paragraph{Same write point, two writers (test game with 12 rooms, $\Fp=20$, step 103, 69 steps after the reveal).}
Recipe: \emph{dice the yellow bell pepper; grill the yellow bell pepper; chop the purple potato; roast the purple potato; slice the pork chop; grill the pork chop}. The \fmt{summary} state (coverage 0.33) reads \emph{In corridor; came from west; goal: dice pepper, chop potatoes, grill both, slice pork chop, roast potato, grill pork chop, then enjoy meal; exits: north (no door), south (available), west (back to kitchen); no items carried}, from which the reader chooses a wrong option ($\NLL(a^*)=2.27$). The trained writer's state (seed 1, coverage 0.50) reads \emph{In bedroom; exit west; previously in kitchen; need pork chop, bell pepper, potatoes; goal: find ingredients, cook meal by grilling pork chop with diced bell pepper and potatoes, then enjoy}, from which the reader picks $a^*$ ($\NLL=0.00$). Both states have paraphrased the directions; the trained state keeps the ingredient list and the current room graph. Ten such same-write-point comparisons per seed are in the released supplement.